\documentclass{article}

     \usepackage[final]{neurips_2023}

\usepackage[utf8]{inputenc} 
\usepackage[T1]{fontenc}    
\usepackage{hyperref}       
\usepackage{url}            
\usepackage{booktabs}       
\usepackage{amsfonts}       
\usepackage{pifont}
\usepackage{amssymb}
\usepackage{nicefrac}       
\usepackage{microtype}      
\usepackage{xcolor}         
\usepackage{graphicx}
\usepackage{multirow}
\usepackage{comment}
\usepackage{mathtools}
\usepackage{graphicx}
\usepackage{caption}
\usepackage{subcaption}
\usepackage{tablefootnote}
\usepackage{textcomp} \usepackage{scalerel}

\newcommand{\cmark}{\ding{51}}%

\usepackage{caption}

\usepackage{hyperref}
\usepackage{url}
\usepackage{graphicx}
\usepackage{booktabs}
\usepackage{algorithm}
\usepackage{algpseudocode}
\usepackage{multirow}
\usepackage{enumitem}
\usepackage{cleveref}
\usepackage{xcolor,colortbl}
\usepackage{wrapfig,lipsum,booktabs}
\usepackage{bbm}
\usepackage[most]{tcolorbox}
\usepackage{lmodern}
\usepackage{tikz}
\usetikzlibrary{arrows.meta} 

\hypersetup{colorlinks, linkcolor={red!55!black}, citecolor={blue!65!black}, urlcolor={blue!65!black}}

\newenvironment{pcrfont}{\fontfamily{pcr}\selectfont}{\par}
\DeclareTextFontCommand{\textpcr}{\pcrfont}

\title{Self-Rewarding Language Models}

\author{%
  Weizhe Yuan$^{1,2}$ \quad Richard Yuanzhe Pang$^{1,2}$  \quad {Kyunghyun Cho}$^{2}$  \\ {\bf Xian Li}$^{1}$ \quad {\bf Sainbayar Sukhbaatar}$^{1}$  \quad     \textbf{Jing Xu}$^{1}$ 
  \quad \textbf{Jason Weston}$^{1,2}$   \\\\
  $^1$ Meta ~~~~~~~~~~ $^{2}$ NYU \\
}

\begin{document}

\maketitle

\begin{abstract}
We posit that to achieve superhuman agents, future models require 
superhuman feedback in order to provide an adequate training signal.
Current approaches commonly train reward models from human preferences, which may then be
bottlenecked by human performance level,
and secondly these separate  frozen reward models  cannot then learn to improve during LLM training.
In this work, we study {\em Self-Rewarding Language Models}, where the language model itself is used via LLM-as-a-Judge prompting to provide its own rewards during training. We show that 
during Iterative DPO training that 
 not only does instruction following ability improve, but also the ability to provide high-quality rewards to itself. 
Fine-tuning Llama 2 70B on three iterations of our approach yields a model that outperforms 
many existing systems on the AlpacaEval 2.0 leaderboard, including Claude 2, Gemini Pro, and GPT-4 0613.
While there is much left still to explore,  this work opens the door to the possibility of models that can continually improve in both axes.
\end{abstract}

\section{Introduction}

Aligning Large Language Models (LLMs) 
using human preference data can vastly improve the instruction following 
performance of pretrained models \citep{ouyang2022training,bai2022training}. 
The standard approach of Reinforcement Learning from Human Feedback (RLHF) 
learns a reward model from these human preferences. The reward model is then frozen and used to train the LLM using RL, e.g., via  PPO \citep{schulman2017proximal}.  A recent alternative is to avoid training the reward model at all, and directly use human preferences to train the LLM, as in Direct Preference Optimization \citep[DPO;][]{rafailov2023direct}.
In both cases, the approach is bottlenecked by the size and quality of the human preference data, and in the case of RLHF the quality of the frozen reward model trained from them as well. 

In this work, we instead
propose to train a self-improving reward model
that, rather than being frozen,  is continually updating during LLM alignment, in order to avoid this bottleneck. 
The key to such an approach is to develop an agent that possesses all the abilities desired during training, rather than separating them out into distinct models such as a reward model and a language model. 
In the same way that pretraining and 
multitasking training of instruction following tasks
allow task transfer by training on many tasks at once 
\citep{collobert2008unified,radford2019language,ouyang2022training}, incorporating the reward model into that same system allows task transfer between the reward modeling task and the instruction following tasks.

We thus introduce {\em Self-Rewarding Language Models}, that both (i) act as instruction following models generating responses for given prompts; and (ii) can generate and evaluate new instruction following examples to add to their own training set. We train these models using an Iterative DPO framework similar to that recently introduced in \citet{xu2023some}. Starting from a seed model,  in each iteration there is a process of {\em Self-Instruction creation} whereby candidate responses are generated by the model for newly created prompts, and  are then assigned rewards by that same model. The latter is implemented via LLM-as-a-Judge prompting, which can also be seen as an instruction following task. A preference dataset is built from the generated data, and the next iteration of the model is trained via DPO, see
 \autoref{fig:model}.

In our experiments,  we start with a Llama 2 70B \citep{touvron2023llama2} seed model fine-tuned on  Open Assistant \citep{kopf2023openassistant}, and then perform the above training scheme.
We find that not only does the instruction following performance improve from Self-Rewarding LLM alignment compared to the baseline seed model, but importantly
the reward modeling ability, which is no longer fixed, improves as well. 
This means that the model during iterative training is able, at a given iteration, to provide a higher quality preference dataset to itself than in the previous iteration.
While this effect likely saturates in real-world settings, it provides the intriguing possibility of obtaining reward models (and hence LLMs) that are superior to ones that could have been trained from the original human-authored seed data alone.

\section{Self-Rewarding Language Models}

Our approach first assumes access to a base pretrained language model, and a small amount of human-annotated seed data.
We then build a model that aims to possess two skills simultaneously:
\begin{enumerate} 
\item {\em Instruction following}: given a prompt that describes a user request, the ability to generate a high quality, helpful (and harmless) response. 
\item {\em Self-Instruction creation}: the ability to generate and evaluate
new instruction-following examples 
to add to its own training set.
\end{enumerate}

These skills are used so that the model can perform self-alignment, i.e., they are the components used to iteratively train itself using AI Feedback (AIF). 

{\em Self-instruction creation} consists of generating candidate responses and then the model itself judging their quality, i.e., it acts as its own reward model, replacing the need for an external one. This is implemented via the  {\em LLM-as-a-Judge} mechanism \citep{zheng2023judging}, i.e.,
by formulating the evaluation of responses as an instruction following task.
This self-created AIF preference data is used as a training set.

\begin{figure*}[t!]
    \centering
    \includegraphics[width=\linewidth]{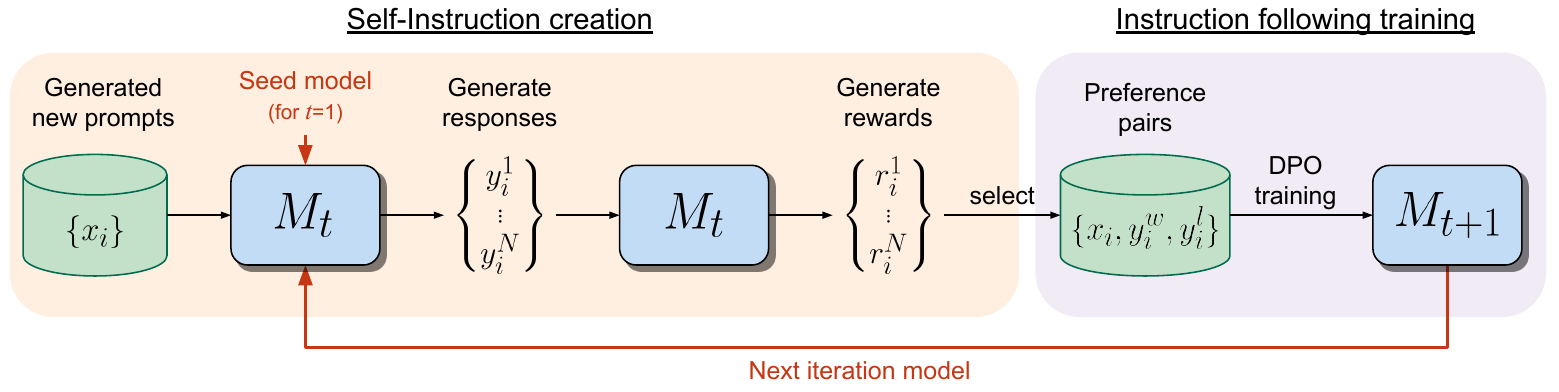}
    \caption{ {\bf Self-Rewarding Language Models.} Our self-alignment method consists of two steps: (i) {\em Self-Instruction creation}:  newly created prompts are used to generate candidate responses from model $M_t$, which also predicts its own rewards via LLM-as-a-Judge prompting. (ii) {\em Instruction following training}: preference pairs are selected from the generated data, 
    which are used for training via DPO, resulting in model $M_{t+1}$. 
    This whole procedure can then be iterated 
    resulting in both improved instruction following and reward modeling ability. }
    \label{fig:model}
\end{figure*}

Our overall {\em self-alignment} procedure is an iterative one, which proceeds by building a series of such models, with the aim that each improves over the last. 
Importantly, because the model can both improve its generation ability, and act as its own reward model through the same generation mechanism, this means the reward model itself can improve through these iterations, deviating from standard practices where the reward model is fixed \citep{ouyang2022training}.
We believe this can increase the ceiling of the potential for self-improvement of 
these learning models going forward, removing a constraining bottleneck.

We describe these steps in more detail below. An overview of the approach is illustrated in Figure \ref{fig:model}.

\subsection{Initialization} \label{sec:init}

\paragraph{Seed instruction following data} We are given a seed set of human-authored 
(instruction prompt, response) general instruction following examples that we use for training in a supervised fine-tuning (SFT) manner, starting from a pretrained base language model. Subsequently this will be referred to as Instruction
Fine-Tuning (IFT) data.

\paragraph{Seed LLM-as-a-Judge instruction following data}
We also assume we are provided a seed set of (evaluation instruction prompt, evaluation result response) examples which can also be used for training. 
While this is not strictly necessary, as the model using IFT data will already be capable of training an LLM-as-a-Judge,  we show that such training data can give 
improved performance (see Appendix~\ref{app:self_reward_ift_only} for supporting results). 
In this data, the input prompt asks the model to evaluate the quality of a given response to a particular instruction. The provided evaluation result response consists of chain-of-thought reasoning (a justification), followed by a final score (in our experiments out of 5). The exact prompt format we chose is given  in \autoref{tab:eval_prompt}, which  instructs the LLM to evaluate the response using five additive criteria (relevance, coverage, usefulness, clarity and expertise), covering various aspects of quality. 
Subsequently this will be referred to as Evaluation Fine-Tuning (EFT) data.




We use  both these seed sets together during training. 


\subsection{Self-Instruction Creation} \label{sec:creation}

Using the model we have trained, we can make it self-modify its own training set.
Specifically, we generate additional training data for the next iteration of training.

This consists of the following steps:

 \begin{enumerate}[topsep=0pt,itemsep=4pt,parsep=0pt] 

\item { Generate a new prompt}: We generate a new prompt $x_i$ using few-shot prompting, sampling prompts from the original seed IFT data, following the approach of \citet{wang2022self} and \citet{honovich2022unnatural}.\footnote{In our main experiments, responses and rewards, items (2) and (3), are generated by the model we have trained, but generating prompts is actually done by a model fixed in advance. However, we show that prompts can also be generated by the newly trained model in each iteration in Appendix~\ref{app:augmented_prompts_generation_using_newly_trained_models}.} 


\item {Generate candidate responses}: We then generate $N$ diverse candidate responses $\{y_i^1, \ldots, y_i^N\}$ for the given prompt $x_i$ from our model using sampling. 

\item {Evaluate candidate responses}: Finally, we use the {LLM-as-a-Judge} ability of our same model to evaluate its own candidate responses with scores $r_i^n \in [0,5]$ (exact prompt given in \autoref{tab:eval_prompt}). 

 \end{enumerate}

\subsection{Instruction Following Training}

As previously described, training is initially performed with the seed IFT and EFT data (\autoref{sec:init}). 
This is then augmented with additional data via AI (Self-)Feedback. 

\paragraph{AI Feedback Training}
After performing the self-instruction creation procedure, we can 
augment the  seed data with  additional examples for training, which we refer to as
AI Feedback Training (AIFT) data. 

To do this, we construct {\em preference pairs}, which are training data of the form (instruction prompt $x_i$, winning response $y_i^w$, losing response $y_i^l$).  To form the winning and losing pair we take the highest and lowest scoring responses from the $N$ evaluated candidate responses (see \autoref{sec:creation}), following \citet{xu2023some}, discarding the pair if their scores are the same. These pairs can be used for training with a preference tuning algorithm. We use DPO \citep{rafailov2023direct}. 

\begin{figure}[t]
\centering
\small
\begin{tcolorbox}[colback=green!10!white, 
                  colframe=green!30!white, 
                  width=0.99\textwidth, 
                  arc=4mm, 
                  auto outer arc,
                  ]
Review the user's question and the corresponding response using the additive 5-point scoring system described below. Points are accumulated based on the satisfaction of each criterion:\\
\\
- Add 1 point if the response is relevant and provides some information related to the user's inquiry, even if it is incomplete or contains some irrelevant content.\\
- Add another point if the response addresses a substantial portion of the user's question, but does not completely resolve the query or provide a direct answer.\\
- Award a third point if the response answers the basic elements of the user's question in a useful way, regardless of whether it seems to have been written by an AI Assistant or if it has elements typically found in blogs or search results.\\
- Grant a fourth point if the response is clearly written from an AI Assistant's perspective, addressing the user's question directly and comprehensively, and is well-organized and helpful, even if there is slight room for improvement in clarity, conciseness or focus.\\
- Bestow a fifth point for a response that is impeccably tailored to the user's question by an AI Assistant, without extraneous information, reflecting expert knowledge, and demonstrating a high-quality, engaging, and insightful answer.\\
\\
\\
User: \texttt{\color{red}<INSTRUCTION\_HERE>}\\
\\
<response>\texttt{\color{red}<RESPONSE\_HERE>}</response>\\
\\
After examining the user's instruction and the response:\\
\\
- Briefly justify your total score, up to 100 words.\\
- Conclude with the score using the format: ``Score: <total points>''\\
\\
Remember to assess from the AI Assistant perspective, utilizing web search knowledge as necessary. To evaluate the response in alignment with this additive scoring model, we'll systematically attribute points based on the outlined criteria.
\end{tcolorbox}
\caption{{\bf LLM-as-a-Judge prompt for our LLM to act as a  reward model} and provide self-rewards for its own model generations. The model is initially trained with seed training data of how to perform well at this task, and then improves at this task further through our self-rewarding  training procedure.}
\label{tab:eval_prompt}
\end{figure}

\subsection{Overall Self-Alignment Algorithm}

\paragraph{Iterative Training}
Our overall procedure trains a series of models $M_1,\dots,M_T$ where each successive model $t$ uses augmented training data created 
by the $t-1^\text{th}$ model.  We thus define AIFT($M_t$) to mean AI Feedback Training data created using model $M_t$. 

\paragraph{Model Sequence}
\label{sec:model-sequence}
We define the  models, and the training data they use as follows: 

\begin{itemize}[topsep=0pt,itemsep=4pt,parsep=0pt]
\item[$M_0$]: Base pretrained LLM with no fine-tuning.

\item[$M_1$]: Initialized with $M_0$, then fine-tuned on the IFT+EFT seed data using SFT.

\item[$M_2$]: Initialized with $M_1$, then trained with AIFT($M_1$) data using DPO.

\item[$M_3$]: Initialized with $M_2$, then trained with AIFT($M_2$) data using DPO.

\end{itemize}

This iterative training resembles the procedure used in Pairwise Cringe Optimization and specifically is termed  Iterative DPO,  introduced in \citet{xu2023some}; however, an external fixed reward model was used in that work.

\section{Experiments}

\subsection{Experimental Setup}

\paragraph{Base Model}
In our experiments we use Llama 2 70B \citep{touvron2023llama2} as our base pretrained model.

\subsubsection{Seed Training Data} \label{sec:seed_data}

\paragraph{IFT Seed Data}
We use the human-authored examples provided in the Open Assistant dataset \citep{kopf2023openassistant} for instruction fine-tuning. Following \citet{li2023self} we use 3200 examples,  by sampling  only first conversational turns in the English language that
are high-quality, based on their human annotated rank (choosing only the highest rank 0).
In our experiments, we compare to a model fine-tuned from the base model using only this data via supervised fine-tuning, and refer to it as our {\em SFT baseline}.

\paragraph{EFT Seed Data} The Open Assistant data also provides multiple ranked human responses per prompt  from which we can construct evaluation fine-tuning data. We split this into train and evaluation sets, and use it to create LLM-as-a-Judge data. This is done by placing it in the input format given in \autoref{tab:eval_prompt}, which consists of the scoring criteria description, and the given instruction and response to be evaluated.\footnote{Note, the prompt, derived from  \citet{li2023self}, mentions ``utilizing web search'', but our model is not actually capable of this action.} For training targets, chain-of-thought justifications and final scores out of 5 are not directly provided, so we use the SFT baseline to generate such output evaluations for each input, 
and accept them into the training set if the ranking of their scores agrees with the human rankings in the dataset. We resample the training set by discarding some of the data that receives the most common score so that the scores are not too skewed, as we observe many samples receive a score of 4. This results in 1,630 train and 541 evaluation examples (which do not overlap with the IFT data).

\subsubsection{Evaluation Metrics}

We evaluate the performance of our self-rewarding models in two axes:
 their ability to follow instructions, and their ability
 as a reward model (ability to evaluate responses).
  
\paragraph{Instruction Following}
We evaluate head-to-head performance between various models using GPT-4 \citep{achiam2023gpt} as an evaluator over 256 test prompts (which we refer to as IFT test data) derived from various sources following \citet{li2023self} using the AlpacaEval evaluation prompt \citep{alpaca_eval}. We try the prompt in both orders comparing pairwise, and if the GPT-4 evaluations disagree we count the result as a tie.
We also perform a similar evaluation with humans (authors).
We additionally report results in the  AlpacaEval 2.0 leaderboard format which is evaluated over 805  prompts, and compute the win rate against the baseline GPT-4 Turbo model based on GPT-4 judgments. Further, we report results on MT-Bench \citep{zheng2023judging} a set of challenging multi-turn questions in various categories from math and coding to roleplay and writing, which uses GPT-4 to grade the model responses out of 10.
Finally we also test the models on a set of 9 NLP benchmarks: ARC-Easy \citep{Clark2018ThinkYH}, ARC-Challenge \citep{Clark2018ThinkYH}, HellaSwag \citep{DBLP:conf/acl/ZellersHBFC19}, SIQA \citep{DBLP:journals/corr/abs-1904-09728}, PIQA \citep{Bisk2020}, GSM8K \citep{cobbe2021gsm8k}, MMLU \citep{DBLP:conf/iclr/HendrycksBBZMSS21}, OBQA \citep{OpenBookQA2018} 
and NQ \citep{47761}. 

\paragraph{Reward Modeling}
We evaluate the correlation with human rankings on the evaluation set we derived from the Open Assistant dataset,  as described in \autoref{sec:seed_data}.
Each instruction has on average 2.85 responses with given rankings. 
We can thus measure the {\em pairwise accuracy}, which is how many times the order of the ranking between any given pair agrees between the model's evaluation and the human ranking.
We also measure the {\em exact match} count, which is how often the total ordering is exactly the same for an instruction. We also report the Spearman correlation and Kendall's $\tau$. 
Finally, we report how often the responses that the model scores a perfect 5 out of 5 are rated as the highest ranked by humans.

\subsubsection{Training Details}

\paragraph{Instruction following training}
The training hyperparameters we use are as follows.
For SFT we use learning rate $5.5e{-6}$ which decays (cosine) to $1.1e{-6}$ at the end of training, batch size $16$ and dropout $0.1$. We only calculate the loss on target tokens instead of the full sequence.
For DPO we use learning rate $1e{-6}$ which decays to $1e{-7}$, batch size $16$, dropout $0.1$, and a $\beta$ value of 0.1.
We perform early stopping by saving a checkpoint every 200 steps and evaluating generations using Claude 2 \citep{claude2} on 253 validation examples derived from various sources following \citet{li2023self}. This is evaluated pairwise against the previous step's generations using the AlpacaEval evaluation prompt format \citep{alpaca_eval}.

\paragraph{Self-Instruction creation}
To generate new prompts we use a fixed model, Llama 2-Chat 70B with 8-shot prompting following Self-Instruct~\citep{wang2022self}, where we sample six demonstrations from the IFT data and two from the model generated data, and use decoding parameters T = 0.6, p = 0.9. We use their prompt template for non-classification tasks and apply the same filtering techniques, including the ROUGE-L~\citep{lin-2004-rouge} similarity check, keyword filtering, and length filtering. Except for the prompt generation part, the other parts of the creation pipeline (generating the response, and evaluating it) use the Self-Rewarding model being trained. For candidate response generation we sample $N=4$ candidate responses with temperature $T = 0.7$, $p=0.9$. 
When evaluating candidate responses, as there is variance to these scores, in our experiments we also use sampled decoding (with the same parameters) and generate these evaluations multiple (3) times and take the average. We added
3,964 such preference pairs to form the AIFT($M_1$) dataset used to train $M_2$ via DPO, 
and 6,942 pairs  to form AIFT($M_2$) used to train $M_3$.

\subsection{Results}
\label{subsec:results}
\subsubsection{Instruction Following Ability}

\begin{figure}[t!]
    \centering
    \includegraphics[width=0.8\textwidth,trim={0 -1cm 0  0}]{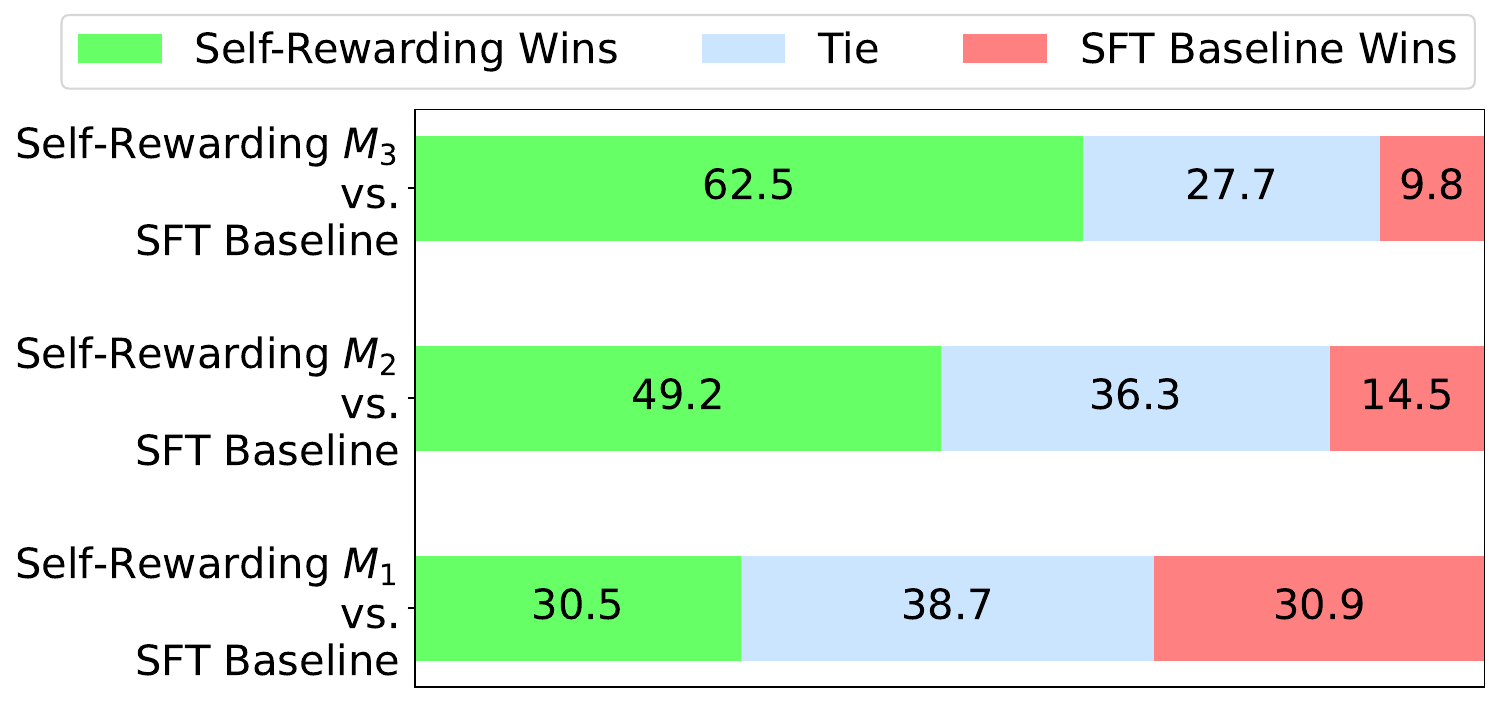}
    \includegraphics[width=0.8\textwidth,trim={0 0 0 0}]{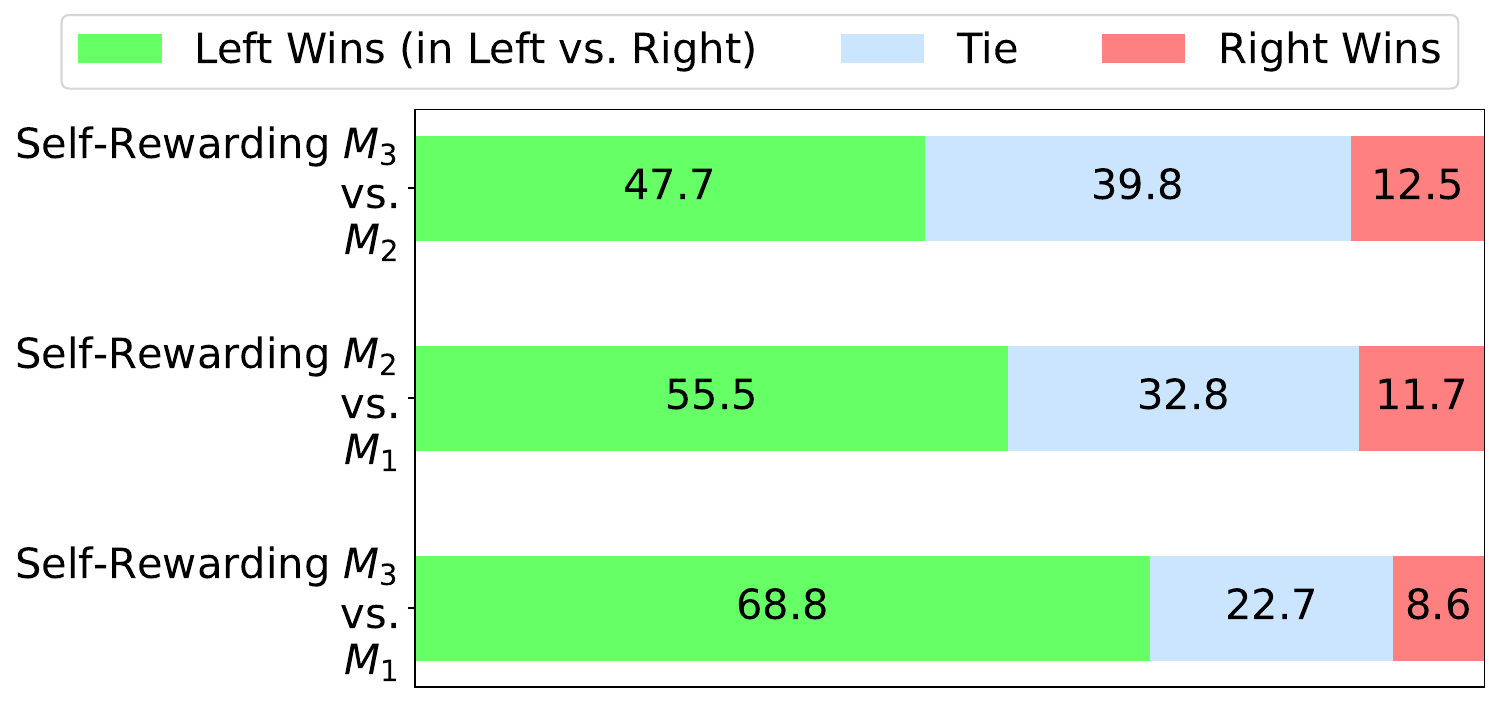}
   \caption{
   \label{fig:h2h}
   {\bf Instruction following ability improves with Self-Training:} 
     We evaluate our models using head-to-head win rates on diverse prompts using GPT-4. %
    The SFT Baseline is on par with Self-Rewarding Iteration 1 ($M_1$).
     However, Iteration 2 ($M_2$) outperforms both Iteration 1 ($M_1$) and the SFT Baseline. 
     Iteration 3 ($M_3$) gives further gains over Iteration 2 ($M_2$), outperforming 
     $M_1$, $M_2$ and the SFT Baseline by a large margin. 
    }
\end{figure}

Head to head performance results are provided in \autoref{fig:h2h}.

\paragraph{EFT+IFT seed training performs similarly to IFT alone}
We find that adding the Evaluation Fine-Tuning (EFT) task to training does not impact instruction following performance compared to using Instruction Fine-Tuning (IFT) data alone
with an almost equal head to head (30.5\% wins vs. 30.9\% wins).
This is a positive result because it means the increased capability of a model to self-reward  does not affect its other skills. We can thus use IFT+EFT training as Iteration 1 ($M_1$) of our Self-Rewarding model, and then run further iterations.

\paragraph{Iteration 2 ($M_2$) improves over Iteration 1 ($M_1$) and SFT Baseline}
Iteration 2 of Self-Rewarding training  ($M_2$) provides superior instruction following to Iteration 1 ($M_1$)  with 55.5\% wins for $M_2$ compared to only 11.7\% for $M_1$ in a head to head evaluation. 
It provides similar gains  over the SFT Baseline as well (49.2\% wins vs. 14.5\% wins). Clearly, there is a large jump in performance from $M_1$ to $M_2$ by using the preference data AIFT($M_1$) provided by the reward model from Iteration 1.

\paragraph{Iteration 3 ($M_3$) improves over Iteration 2 ($M_2$)}
We see a further gain in Iteration 3 over Iteration 2, with 47.7\% wins for $M_3$ compared to only 12.5\% for $M_2$ in a head to head evaluation. Similarly, the win rate over the SFT Baseline for $M_3$ increases to 62.5\% wins vs. 9.8\%, i.e., winning more often than the $M_2$ model did. Overall, we see large gains from $M_2$ to $M_3$ through training using the preference data AIFT($M_2$) provided by the reward model from Iteration 2.

\paragraph{Self-Rewarding models perform well on AlpacaEval 2 leaderboard}
We evaluate our models on the AlpacaEval 2.0 leaderboard format, with results given in 
\autoref{tab:alpaca_leaderb}. We observe the same findings as in the head-to-head evaluations, that training iterations yield improved win rates, in this case over GPT4-Turbo, from 9.94\% in Iteration 1, to 15.38\% in Iteration 2, to 20.44\% in Iteration 3. Our Iteration 3 model outperforms many existing models in this metric, including Claude 2, Gemini Pro, and GPT4 0613. We show some selected models from the leaderboard in the table. We note that many of those competing models contain either proprietary alignment data (which is typically large, e.g., over 1M annotations in \citet{touvron2023llama2}) or use targets that are distilled from stronger models. In contrast, our Self-Rewarding model  starts from a small set of seed data from Open Assistant, and then generates targets and rewards from the model itself for further iterations of  training. 

\begin{table}[h]
    \caption{
    {\bf AlpacaEval 2.0 results} (win rate over GPT-4 Turbo evaluated by GPT-4). Self-Rewarding iterations yield improving  win rates. Iteration 3 ($M_3$) outperforms many existing models that use proprietary training data or targets distilled from stronger models.
    }
    \vspace{2mm}
  \label{tab:alpaca_leaderb}
  \centering

  \begin{tabular}{llcc}
    \toprule
                   &                          & \multicolumn{2}{c}{ Alignment Targets} \\
    \textbf{Model} &    \textbf{Win Rate}  & Distilled & Proprietary\\
    \midrule  
    Self-Rewarding 70B \\
    ~~~~ {\em Iteration 1} ($M_1$) & 9.94\% & \\ 
    ~~~~ {\em Iteration 2} ($M_2$) & 15.38\%&   \\ 
    ~~~~ {\em Iteration 3} ($M_3$) & 20.44\% & \\ 
    \midrule  
     \midrule 
   {\em Selected models from the leaderboard}\\
   GPT-4 0314	& 22.07\% &   & \cmark  \\ 
   Mistral Medium & 	21.86\% &   & \cmark   \\ 
   Claude 2 &	17.19\%&   & \cmark  	\\ 
   Gemini Pro &	16.85\%  &   & \cmark  \\	
   GPT-4 0613 &	15.76\% &   & \cmark  	\\	
   GPT 3.5 Turbo 0613	&14.13\%  &   & \cmark  \\ 
   LLaMA2 Chat 70B &	13.87\% &   & \cmark   \\ 
   Vicuna 33B v1.3 &	12.71\% &   \cmark & \\ 
   Humpback LLaMa2 70B &	10.12\% \\ 
   Guanaco 65B &	6.86\% & & \\ 
   Davinci001 &	2.76\%  &  & \cmark \\ 
   Alpaca 7B &	2.59\%  &   \cmark &   \\
    \bottomrule
  \end{tabular}
\end{table}

\begin{figure}[h!]
    \centering
    \begin{subfigure}[b]{0.7\linewidth}
        \includegraphics[width=\columnwidth]{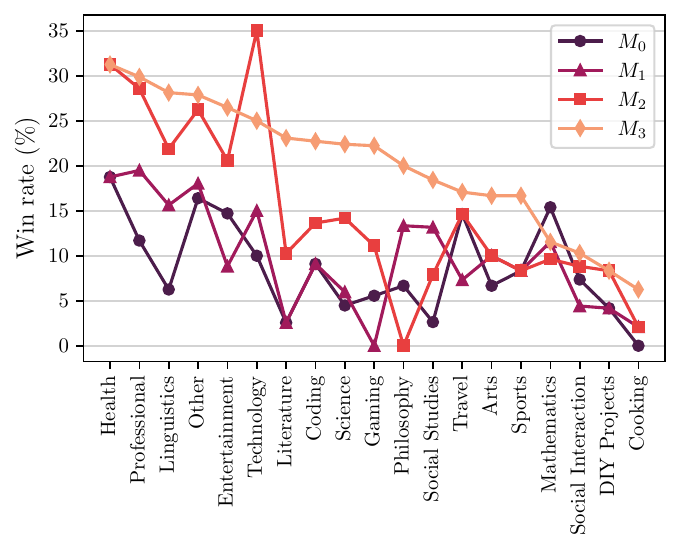}
        \label{fig:topic_winrate}
    \end{subfigure}
    \captionsetup{skip=-10pt}  
    \caption{AlpacaEval win rate breakdown for instruction categories (full names given in Appendix).
    Self-Rewarding models give gains across several topics, but tend to e.g. give less gains on mathematics and reasoning tasks.
    }
    \label{fig:fine_grained_winrate}
\end{figure}


\paragraph{Fine-grained analysis}
As described earlier, the overall performance of the model in AlpacaEval improves with each iteration of training. It would be interesting to break down the overall performance improvement to see exactly what type of tasks these improvements come from. Therefore, we cluster the instructions in AlpacaEval test set into different groups based on three perspectives: (1) instruction category (2) instruction complexity (3) expected response length. We achieve this by using GPT-4. The detailed statistical information of the breakdown and the prompting techniques we used for getting this breakdown can be found in Appendix~\ref{app:alpacaeval_test_sample_clustering}. Results for  the instruction category are given in \autoref{fig:fine_grained_winrate}, and the other two in Appendix \autoref{fig:fine_grained_winrate_detailed_other_two}. From the results we can conclude that (i) Self-Rewarding models can substantially improve the win rate in most categories, but there are some tasks for which this approach does not improve, such as mathematics and logical reasoning, indicating that our current training approach mainly allows the models to better utilize their existing knowledge.
(ii) Through Self-Rewarding model training, the model's win rate increases on almost all tasks of different complexity, and especially on slightly more difficult tasks (complexity of 5, 6, 7 out of 10). (iii) The models also show a steady increase in the win rate on tasks with instructions with different expected response lengths.

\paragraph{Data distribution analysis}
We perform a t-SNE \citep{van2008visualizing} visualization of the IFT, EFT and AIFT($M_1$) data, shown in Appendix~\ref{app:ift_eft_distribution}. We find good overlap between the IFT and AIFT($M_1$) examples, which is desired, while the EFT examples lie in a different part of the embedding space, which can help explain why they would not affect IFT performance. 
We observe that generations from $M_1$ on AlpacaEval have an average length of 1092, for $M_2$ they are 
1552, and for $M_3$ they are 2552, so the model is learning to generate longer responses, which we note may be a factor in relative performance.

\begin{figure}[t!]
    \centering
    \includegraphics[width=0.8\textwidth,trim={0 0 0 0}]{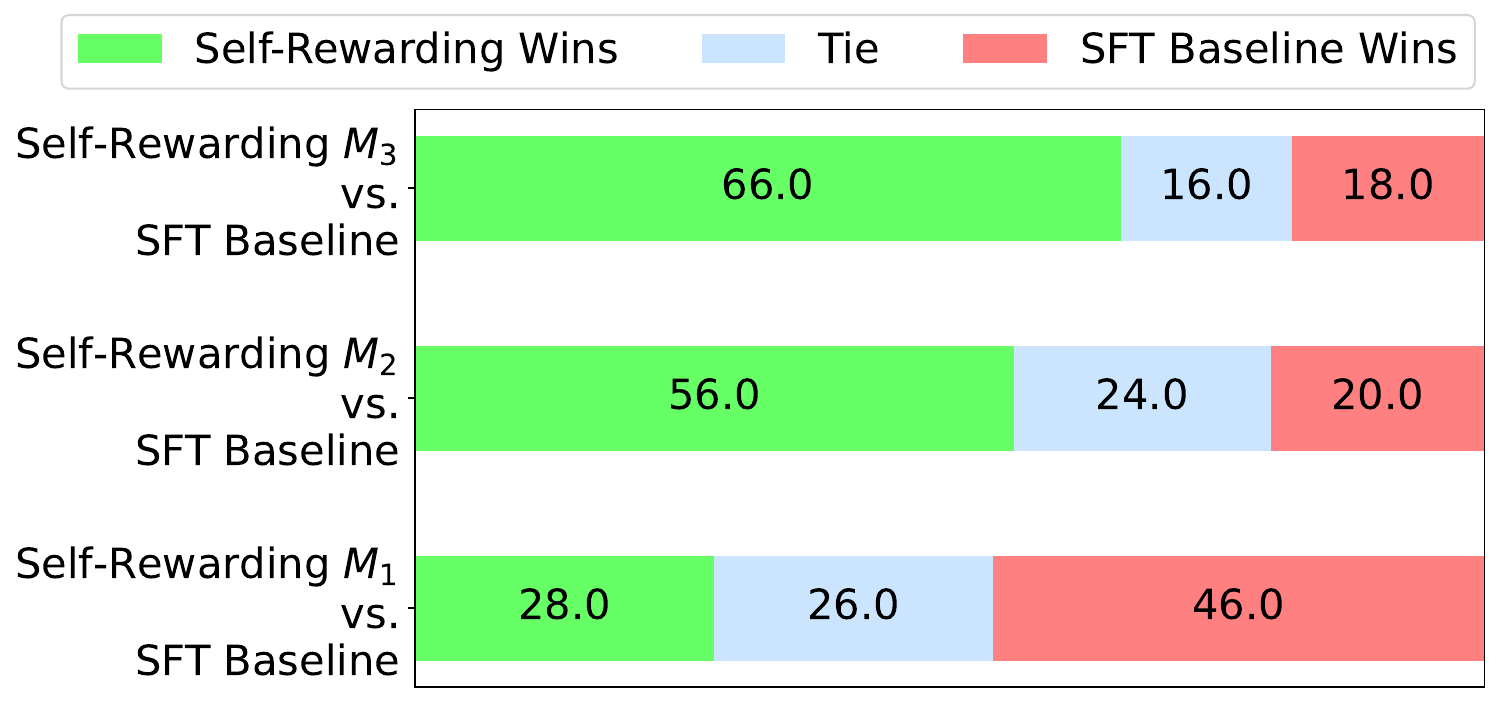}
   \caption{
   \label{fig:h2h_humaneval}
   {\bf } 
     {\bf Human evaluation results.} Iterations of Self-Rewarding ($M_1$, $M_2$ and $M_3$) provide progressively better head-to-head win rates compared to the SFT baseline, in agreement with  the automatic evaluation results. 
    }

\end{figure}

\begin{table}[h]
    \caption{{\bf MT-Bench Results} (on a scale of 10). Self-Rewarding iterations yield improving scores across various categories. Math, code \& reasoning performance and iteration gains are smaller than for other categories, likely due to the makeup of the Open Assistant seed data we use.}
  \label{tab:mt-bench}
  \centering
\begin{tabular}{lccc}
\toprule
               &  Overall         & Math, Code &  Humanities, Extraction,  \\
               &  Score & \& Reasoning &  STEM, Roleplay \& Writing \\
                \midrule
{SFT Baseline}    & 6.85 & 3.93 & 8.60\\
{$M_1$}  & 6.78 & 3.83 & 8.55\\
{$M_2$}  & 7.01 & 4.05 & 8.79    \\                                                       
{$M_3$}   & 7.25 & 4.17 & 9.10  \\  
\bottomrule
\end{tabular}
\end{table}

\begin{table}[h!]
    \caption{
    {{\bf NLP Benchmarks.} Self-Rewarding models mostly tend to maintain performance compared to the  Llama 2 70B base model and the SFT Baseline, despite being fine-tuned on very different instruction-following prompts.}
    }
  \label{tab:core_knowledge_eval_results}
  \centering
\begin{tabular}{lccccc}
\toprule
               & \multicolumn{1}{c}{{\begin{tabular}[c]{@{}c@{}}ARC $(\uparrow)$\\challenge\end{tabular}}}
               & \multicolumn{1}{c}{{\begin{tabular}[c]{@{}c@{}}HellaSwag\\$(\uparrow)$\end{tabular}}}
               & \multicolumn{1}{c}{{\begin{tabular}[c]{@{}c@{}}GSM8K\\$(\uparrow)$\end{tabular}}}
               & \multicolumn{1}{c}{{\begin{tabular}[c]{@{}c@{}}MMLU\\$(\uparrow)$\end{tabular}}}
              & \multicolumn{1}{c}{{\begin{tabular}[c]{@{}c@{}}NQ\\$(\uparrow)$\end{tabular}}} \\
                \midrule
{Llama 2}                                  & 57.40                                        & 85.30                                                             & 56.80                                                                              & 68.90                                                                                                                                                                     & 25.30                                                                           \\
{SFT Baseline}                                   & 55.97                                       & 85.17                                                          & 50.72                                                                             & 69.76                                                                                                                                                           & 34.35                                                                          \\
{$M_1$}                                    & 57.51                                       & 84.99                                                           & 60.27                                                                             & 69.34                                                                                                                                                         & 35.48                                                                          \\
{$M_2$}                                  & 54.51                                       & 84.27                                                    & 59.29                                                                             & 69.31                                                                                                                                                               & 33.07                                                                          \\
{$M_3$}                                  & 53.13                                       & 83.29                                                      & 57.70                                                                              & 69.37                                                                                                                                                                & 31.86  \\
\bottomrule
\end{tabular}
\end{table}

\paragraph{Human evaluation}  
To examine whether human judgments align with automatic evaluation results, we conduct human evaluations that compare SFT baseline generations with the generations from each iteration of Self-Rewarding training, i.e., models $M_1$, $M_2$, and $M_3$. Specifically, we randomly select 50 instructions from the IFT test set. Each instruction corresponds to three pairs of generations (i.e., baseline vs. $M_1$, baseline vs. $M_2$, baseline vs. $M_3$). For each pair of generations, we assign them to three different annotators (blind evaluation performed by the authors) to make a pairwise judgment, and take a majority vote to decide which generation is better. The human evaluation results are shown in \autoref{fig:h2h_humaneval}. We find that Self-Rewarding models from later iterations show a larger advantage over the SFT baseline model, which is consistent with GPT-4's judgments, and demonstrates the effectiveness of our iterative training procedure.

\paragraph{MT-Bench performance further validates these results}  
We report performance on MT-Bench in \autoref{tab:mt-bench} for the SFT baseline and iterations of the Self-Rewarding model.  We again see improvements across the iterations of training from $M_1$ to $M_3$, from 6.78 (out of 10) up to 7.25, with larger relative gains in the humanities, STEM, roleplay, writing and extraction categories, and smaller gains in the math, code and reasoning categories.
We expect that the latter is due to the seed prompts we use from Open Assistant tending to underemphasize the reasoning-based tasks.
We note also that these improvements are in spite of our method using and constructing prompts that only involve a single turn, given the MT-Bench benchmark itself is a multi-turn evaluation.

\paragraph{Self-rewarding models did not lose ability on NLP Benchmarks} 
As shown in \autoref{tab:core_knowledge_eval_results}, the performance of most NLP benchmark tasks evaluated are roughly similar to the baselines, with further detailed results on more datasets given in Appendix \autoref{tab:core_knowledge_eval_results_detailed}  that follow the same pattern.
 We hypothesize that given that our training data (seed data and synthetically generated data) are based on the Open Assistant prompts which may not be especially relevant to skills needed in the \autoref{tab:core_knowledge_eval_results} tasks, it is expected that the task performance stays roughly similar, or may even drop. 
 For example, in InstructGPT training \citep{ouyang2022training}
 they found that ``during RLHF fine-tuning, we observe performance regressions compared
to GPT-3 on certain public NLP datasets'' which they refer to as an ``alignment tax.'' 
 A clear future direction is to extend the self-rewarding paradigm to these types of tasks, by relying not only on seed prompts from Open Assistant, but also on seed prompts found in a larger variety of datasets.

\subsubsection{Reward Modeling Ability}

Reward modeling evaluation results are provided in 
\autoref{tab:reward_perf}.

\paragraph{EFT augmentation improves over SFT baseline}
Firstly, we find that adding Evaluation Fine-Tuning (EFT) data into training, which gives examples to the model of how to act as an LLM-as-a-Judge, naturally improves its performance compared to training with Instruction Fine-Tuning (IFT) data alone. IFT data covers a wide range of general instruction tasks, and so does endow the SFT Baseline with the ability to evaluate responses; however, EFT data gives more examples of this specific task.  We find improvements across all five metrics measured when using IFT+EFT vs. IFT alone, e.g., 
the pairwise accuracy agreement with humans increases from 65.1\% to 78.7\%. 

\begin{table}[t]
\caption{{\bf Reward Modeling ability improves with Self-Training}:  We evaluate the LLM-as-a-Judge via various metrics which measure alignment with held-out human preference data. Self-Rewarding Iteration 2 (Model $M_2$), which is trained using the self-reward model derived from its previous iteration $M_1$ outperforms Iteration 1 ($M_1$), while $M_1$ itself outperforms a standard SFT baseline model trained on only Instruction Fine-Tuning (IFT) data. Iteration 3 (Model $M_3$) gives further improvements over Iteration 2.
}
\label{tab:reward_perf}
\vspace{2mm}
\begin{center}

\begin{tabular}{lcccc}
\toprule
                  &                    &  \multicolumn{3}{c}{Self-Rewarding Models}   \\ 
Model             &      SFT Baseline  &  Iter 1 ($M_1$) & Iter 2 ($M_2$)  & Iter 3 ($M_3$) \\
\midrule
{Training data}  &       {\footnotesize{IFT}}          &    {\footnotesize{IFT+EFT}}     &    {\footnotesize{IFT+EFT}}      &   {\footnotesize{IFT+EFT}{+AIFT($M_1$)}} \\
              &                    &               &   {\footnotesize {+AIFT($M_1$)}} &  {\footnotesize{+AIFT($M_2$)}}  \\
\midrule
Pairwise acc. $(\uparrow)$ & 65.1\% & 78.7\% & 80.4\% & 81.7\% \\
5-best \%    $(\uparrow)$ &   39.6\% & 41.5\% & 44.3\% & 43.2\% \\
Exact Match \%  $(\uparrow)$ &  10.1\% & 13.1\% & 14.3\% & 14.3\%  \\
Spearman corr. $(\uparrow)$ &  0.253 & 0.279 & 0.331 & 0.349 \\
Kendall $\tau$ corr.    $(\uparrow)$ &   0.233 & 0.253 & 0.315 & 0.324  \\
\bottomrule
\end{tabular}
\end{center}
\end{table}

\paragraph{Reward Modeling ability improves with Self-Training}
We find that performing a round of self-reward training {\em improves the ability of the model at providing self-rewards for the next iteration}, in addition to its improved instruction following ability. Model $M_2$ (Iteration 2) is trained using the reward model from $M_1$ (Iteration 1), but provides improved performance on all five metrics compared to $M_1$.
For example,  pairwise accuracy  improves from 78.7\% to 80.4\%.
Iteration 3 ($M_3$) improves several of these metrics further compared to $M_2$, for example pairwise accuracy increases from 80.4\% to 81.7\%.
This performance gain is achieved despite there being no additional 
EFT data provided, and the examples created during the {\em Self-Instruction creation} loop do not tend to look like LLM-as-a-Judge training examples. We hypothesize that because the model is becoming better at general instruction following, it nevertheless also improves at
the  LLM-as-a-Judge task.

\paragraph{Importance of the LLM-as-a-Judge Prompt}
In these experiments we used the LLM-as-a-Judge prompt format shown  in  \autoref{tab:eval_prompt}.
In preliminary experiments we also tried various other prompts to decide the most effective one to use. For example, we tried the prompt proposed in \citet{li2023self} which also proposes a 5-point scale, but describes the options as multiple choice in a range of quality buckets, see Appendix \autoref{tab:eval_prompt_xian}.
 In contrast, our prompt describes the points as additive, covering various aspects of quality. We find a large difference between these two prompts when using the SFT Baseline, e.g. 65.1\% pairwise accuracy for ours, and only 26.6\% pairwise accuracy for theirs. See Appendix~\ref{app:xian_eft_prompt} for further details.

\section{Related Work}

Automatically improving or self-correcting large language models is becoming
a major focus of research. A recent survey from \citet{pan2023automatically}
attempts to summarize the topic. However, this is a rapidly moving area, and 
there are already promising new works not covered there.

\paragraph{Reinforcement Learning from Human Feedback (RLHF)}
Preference learning approaches such as in \citet{ziegler2019fine,stiennon2020learning,ouyang2022training,bai2022training}
train a fixed reward model from human preference data, and then use the reward model to train via reinforcement learning (RL), e.g. via Proximal Policy Optimization (PPO) \citep{schulman2017proximal}. Thus, the reward signal  in a certain sense  already comes from a model even in these works, but distilled from human data. Nevertheless, this is  commonly referred to as RL from Human Feedback (RLHF). 
Methods such as Direct Preference Optimization (DPO) \citep{rafailov2023direct} avoid training the reward model 
entirely, and instead directly train the LLM using human preferences. Several other such competing methods exist as well \citep{zhao2023slic,zheng2023click,yuan2023rrhf}, including Pairwise Cringe Optimization (PCO) \citep{xu2023some}. 
PCO uses an iterative training approach similar to the one in our work, except with a fixed reward model, and that work also showed that Iterative DPO improves over DPO using the same scheme.
We note that other works have developed iterative preference training schemes
as well, e.g. \citet{adolphs2022cringe,gulcehre2023reinforced,xiong2023gibbs}.

\paragraph{Reinforcement Learning from AI Feedback (RLAIF)}
Constitutional AI \citep{bai2022constitutional}
uses an LLM to give feedback and refine responses, and uses this data to train a reward model. This fixed, separate reward model is then used  to train the language model via RL, called  ``RL from AI Feedback'' (RLAIF). 
\citet{lee2023rlaif} compare RLAIF and RLHF procedures and find the methods they compare perform roughly equally. They
use an  ``off-the-shelf'' LLM to perform LLM-as-a-Judge prompting to build a training set to train a fixed reward model, which is then used for RL training. They also experiment with using the fixed but separate LLM-as-a-Judge model directly, which the authors report is computationally expensive due to using it within PPO training (rather than the offline step in the iterative approach we use in our work, which is relatively computationally cheap).
Finally, SPIN \citep{chen2024self} recently showed they can avoid reward models entirely in an Iterative DPO-like framework by using human labels as the winning response in a pair, and the last iteration's generations as the losing response in the pair. The authors note this has the limitation that once the model generations reach human performance, they are bottlenecked. Further, each input prompt is required to have a human annotated response, in contrast to our work.

\paragraph{Improving LLMs via data augmentation (and curation)}
Several methods have improved LLMs by (self-)creating training data to augment fine-tuning. Self-Instruct \citep{wang2022self} is a method for self-instruction creation of prompts and responses, which can be used to improve a base LLM. We make use of a similar technique in our work, and then use our self-reward model to score them.
Several approaches have also created training data by
distilling from powerful LLMs, and shown 
a weaker LLM can then perform well. For example,  Alpaca \citep{alpaca} fine-tuned a Llama 7B  model with text-davinci-003 instructions created in the style of self-instruct.  Alpagasus \citep{chen2023alpagasus} employed a strong LLM-as-a-Judge (ChatGPT) to curate the Alpaca dataset and filter to a smaller set, obtaining improved results. Instruction Backtranslation \citep{li2023self} similarly augments and curates training data, but augmenting via backtranslating from web documents to predict prompts. The curation is done by the LLM(-as-a-Judge) itself, so can be seen as an instance of a self-rewarding model, but in a specialized setting.
Reinforced Self-Training (ReST) \citep{gulcehre2023reinforced} uses a fixed, external reward to curate new high-quality examples to iteratively add to the training set, improving performance. In our experiments, we found that adding only positive examples in a related manner did not help, whereas  preference pairs did help (see Appendix \autoref{app:positive_examples} for details).

\paragraph{LLM-as-a-Judge}
Using LLM-as-a-Judge prompting to evaluate language models has become a 
standard approach \citep{dubois2023alpacafarm,alpaca_eval,fernandes2023devil,bai2023benchmarking,saha2023branch}, and is being used to train reward models or curate data as well, as described above \citep{lee2023rlaif,chen2023alpagasus,li2023self}. While some works such as \citet{kim2023prometheus} create training data to train an LLM to perform well as a judge, to our knowledge it is not common to combine this training with general instruction following skills as in our work.

\section{Conclusion}

We have introduced Self-Rewarding Language Models, models capable of self-alignment via judging and  training on their own generations. The method learns in an iterative manner, where in each iteration the model creates its  own preference-based instruction training data. This is done by assigning rewards to its own generations via LLM-as-a-Judge prompting, and using Iterative DPO to train on the preferences. We showed that  this training both improves the instruction following capability of the model, as well as its reward-modeling ability across the iterations. While there are many avenues left unexplored, we believe this is exciting  because 
this means  the model is better able to assign rewards in future iterations for improving instruction following -- a kind of virtuous circle. 
While this improvement likely saturates in realistic scenarios,
it still allows for the possibility of continual improvement beyond the human preferences that are typically used to build reward models and instruction following models today.

\section{Limitations}

While we have obtained  promising experimental results, we currently consider them preliminary because there are many avenues yet to explore, among them the topics of further evaluation, including safety evaluation, and understanding the limits of iterative training. 

We showed that the iterations of training improve both instruction following and reward modeling ability, but only ran three iterations in a single setting. A clear line of further research is to understand the ``scaling laws’’ of this effect both for more iterations, and with different language models with more or less capabilities in different settings.

We observed an increase in length in model generations, and there is a known correlation between length and estimated quality, which is a topic that should be understood more deeply in general, and in our results in particular as well. It would also be good to understand if so-called ``reward-hacking’’ can happen within our framework, and in what circumstances. As we are using both a language model as the training reward, and a language model for final evaluation (GPT-4) in some of our benchmarks, even if they are different models, this may require a deeper analysis than we have provided. While the human evaluation we conducted did provide validation of the automatic results, further study could bring more insights.

Another clear further avenue of study is to conduct safety evaluations -- and to explore safety training within our framework.
Reward models have been built exclusively for safety in existing systems \citep{touvron2023llama2}, and a promising avenue here would be to use the LLM-as-a-Judge procedure to evaluate for safety specifically in our self-rewarding training process. 
Given that we have shown that reward modeling ability  improves over training iterations, this could mean that the safety of the model could potentially improve over time as well, with later iterations being able to catch and mitigate more challenging safety situations that earlier iterations cannot.

\newpage

\bibliography{sample}

\begin{thebibliography}{45}
\providecommand{\natexlab}[1]{#1}
\providecommand{\url}[1]{\texttt{#1}}
\expandafter\ifx\csname urlstyle\endcsname\relax
  \providecommand{\doi}[1]{doi: #1}\else
  \providecommand{\doi}{doi: \begingroup \urlstyle{rm}\Url}\fi

\bibitem[Achiam et~al.(2023)Achiam, Adler, Agarwal, Ahmad, Akkaya, Aleman, Almeida, Altenschmidt, Altman, Anadkat, et~al.]{achiam2023gpt}
Josh Achiam, Steven Adler, Sandhini Agarwal, Lama Ahmad, Ilge Akkaya, Florencia~Leoni Aleman, Diogo Almeida, Janko Altenschmidt, Sam Altman, Shyamal Anadkat, et~al.
\newblock {GPT-4} technical report.
\newblock \emph{arXiv preprint arXiv:2303.08774}, 2023.

\bibitem[Adolphs et~al.(2023)Adolphs, Gao, Xu, Shuster, Sukhbaatar, and Weston]{adolphs2022cringe}
Leonard Adolphs, Tianyu Gao, Jing Xu, Kurt Shuster, Sainbayar Sukhbaatar, and Jason Weston.
\newblock The {CRINGE} loss: Learning what language not to model.
\newblock In Anna Rogers, Jordan Boyd-Graber, and Naoaki Okazaki, editors, \emph{Proceedings of the 61st Annual Meeting of the Association for Computational Linguistics (Volume 1: Long Papers)}, pages 8854--8874, Toronto, Canada, July 2023. Association for Computational Linguistics.
\newblock \doi{10.18653/v1/2023.acl-long.493}.
\newblock URL \url{https://aclanthology.org/2023.acl-long.493}.

\bibitem[Anthropic(2023)]{claude2}
Anthropic.
\newblock Claude 2.
\newblock \url{https://www.anthropic.com/index/claude-2}, 2023.

\bibitem[Bai et~al.(2022{\natexlab{a}})Bai, Jones, Ndousse, Askell, Chen, DasSarma, Drain, Fort, Ganguli, Henighan, et~al.]{bai2022training}
Yuntao Bai, Andy Jones, Kamal Ndousse, Amanda Askell, Anna Chen, Nova DasSarma, Dawn Drain, Stanislav Fort, Deep Ganguli, Tom Henighan, et~al.
\newblock Training a helpful and harmless assistant with reinforcement learning from human feedback.
\newblock \emph{arXiv preprint arXiv:2204.05862}, 2022{\natexlab{a}}.

\bibitem[Bai et~al.(2022{\natexlab{b}})Bai, Kadavath, Kundu, Askell, Kernion, Jones, Chen, Goldie, Mirhoseini, McKinnon, et~al.]{bai2022constitutional}
Yuntao Bai, Saurav Kadavath, Sandipan Kundu, Amanda Askell, Jackson Kernion, Andy Jones, Anna Chen, Anna Goldie, Azalia Mirhoseini, Cameron McKinnon, et~al.
\newblock Constitutional {AI}: Harmlessness from {AI} feedback.
\newblock \emph{arXiv preprint arXiv:2212.08073}, 2022{\natexlab{b}}.

\bibitem[Bai et~al.(2023)Bai, Ying, Cao, Lv, He, Wang, Yu, Zeng, Xiao, Lyu, Zhang, Li, and Hou]{bai2023benchmarking}
Yushi Bai, Jiahao Ying, Yixin Cao, Xin Lv, Yuze He, Xiaozhi Wang, Jifan Yu, Kaisheng Zeng, Yijia Xiao, Haozhe Lyu, Jiayin Zhang, Juanzi Li, and Lei Hou.
\newblock Benchmarking foundation models with language-model-as-an-examiner.
\newblock In \emph{Thirty-seventh Conference on Neural Information Processing Systems Datasets and Benchmarks Track}, 2023.
\newblock URL \url{https://openreview.net/forum?id=IiRHQ7gvnq}.

\bibitem[Bisk et~al.(2020)Bisk, Zellers, Bras, Gao, and Choi]{Bisk2020}
Yonatan Bisk, Rowan Zellers, Ronan~Le Bras, Jianfeng Gao, and Yejin Choi.
\newblock Piqa: Reasoning about physical commonsense in natural language.
\newblock In \emph{Thirty-Fourth AAAI Conference on Artificial Intelligence}, 2020.

\bibitem[Chen et~al.(2024{\natexlab{a}})Chen, Li, Yan, Wang, Gunaratna, Yadav, Tang, Srinivasan, Zhou, Huang, et~al.]{chen2023alpagasus}
Lichang Chen, Shiyang Li, Jun Yan, Hai Wang, Kalpa Gunaratna, Vikas Yadav, Zheng Tang, Vijay Srinivasan, Tianyi Zhou, Heng Huang, et~al.
\newblock {AlpaGasus}: Training a better alpaca with fewer data.
\newblock In \emph{The Twelfth International Conference on Learning Representations}, 2024{\natexlab{a}}.
\newblock URL \url{https://openreview.net/forum?id=FdVXgSJhvz}.

\bibitem[Chen et~al.(2024{\natexlab{b}})Chen, Deng, Yuan, Ji, and Gu]{chen2024self}
Zixiang Chen, Yihe Deng, Huizhuo Yuan, Kaixuan Ji, and Quanquan Gu.
\newblock Self-play fine-tuning converts weak language models to strong language models.
\newblock \emph{arXiv preprint arXiv:2401.01335}, 2024{\natexlab{b}}.

\bibitem[Clark et~al.(2018)Clark, Cowhey, Etzioni, Khot, Sabharwal, Schoenick, and Tafjord]{Clark2018ThinkYH}
Peter Clark, Isaac Cowhey, Oren Etzioni, Tushar Khot, Ashish Sabharwal, Carissa Schoenick, and Oyvind Tafjord.
\newblock Think you have solved question answering? {T}ry {ARC}, the {AI2} reasoning challenge.
\newblock \emph{arXiv preprint arXiv:1803.05457}, 2018.

\bibitem[Cobbe et~al.(2021)Cobbe, Kosaraju, Bavarian, Chen, Jun, Kaiser, Plappert, Tworek, Hilton, Nakano, Hesse, and Schulman]{cobbe2021gsm8k}
Karl Cobbe, Vineet Kosaraju, Mohammad Bavarian, Mark Chen, Heewoo Jun, Lukasz Kaiser, Matthias Plappert, Jerry Tworek, Jacob Hilton, Reiichiro Nakano, Christopher Hesse, and John Schulman.
\newblock Training verifiers to solve math word problems.
\newblock \emph{arXiv preprint arXiv:2110.14168}, 2021.

\bibitem[Collobert and Weston(2008)]{collobert2008unified}
Ronan Collobert and Jason Weston.
\newblock A unified architecture for natural language processing: Deep neural networks with multitask learning.
\newblock In \emph{Proceedings of the 25th International Conference on Machine Learning}, pages 160--167, 2008.

\bibitem[Dubois et~al.(2023)Dubois, Li, Taori, Zhang, Gulrajani, Ba, Guestrin, Liang, and Hashimoto]{dubois2023alpacafarm}
Yann Dubois, Xuechen Li, Rohan Taori, Tianyi Zhang, Ishaan Gulrajani, Jimmy Ba, Carlos Guestrin, Percy Liang, and Tatsunori~B Hashimoto.
\newblock Alpacafarm: A simulation framework for methods that learn from human feedback.
\newblock \emph{arXiv preprint arXiv:2305.14387}, 2023.

\bibitem[Fernandes et~al.(2023)Fernandes, Deutsch, Finkelstein, Riley, Martins, Neubig, Garg, Clark, Freitag, and Firat]{fernandes2023devil}
Patrick Fernandes, Daniel Deutsch, Mara Finkelstein, Parker Riley, Andr{\'e} Martins, Graham Neubig, Ankush Garg, Jonathan Clark, Markus Freitag, and Orhan Firat.
\newblock The devil is in the errors: Leveraging large language models for fine-grained machine translation evaluation.
\newblock In Philipp Koehn, Barry Haddow, Tom Kocmi, and Christof Monz, editors, \emph{Proceedings of the Eighth Conference on Machine Translation}, pages 1066--1083, Singapore, December 2023. Association for Computational Linguistics.
\newblock \doi{10.18653/v1/2023.wmt-1.100}.
\newblock URL \url{https://aclanthology.org/2023.wmt-1.100}.

\bibitem[Gulcehre et~al.(2023)Gulcehre, Paine, Srinivasan, Konyushkova, Weerts, Sharma, Siddhant, Ahern, Wang, Gu, et~al.]{gulcehre2023reinforced}
Caglar Gulcehre, Tom~Le Paine, Srivatsan Srinivasan, Ksenia Konyushkova, Lotte Weerts, Abhishek Sharma, Aditya Siddhant, Alex Ahern, Miaosen Wang, Chenjie Gu, et~al.
\newblock Reinforced self-training (rest) for language modeling.
\newblock \emph{arXiv preprint arXiv:2308.08998}, 2023.

\bibitem[Hendrycks et~al.(2021)Hendrycks, Burns, Basart, Zou, Mazeika, Song, and Steinhardt]{DBLP:conf/iclr/HendrycksBBZMSS21}
Dan Hendrycks, Collin Burns, Steven Basart, Andy Zou, Mantas Mazeika, Dawn Song, and Jacob Steinhardt.
\newblock Measuring massive multitask language understanding.
\newblock In \emph{9th International Conference on Learning Representations, {ICLR} 2021, Virtual Event, Austria, May 3-7, 2021}. OpenReview.net, 2021.
\newblock URL \url{https://openreview.net/forum?id=d7KBjmI3GmQ}.

\bibitem[Honovich et~al.(2023)Honovich, Scialom, Levy, and Schick]{honovich2022unnatural}
Or~Honovich, Thomas Scialom, Omer Levy, and Timo Schick.
\newblock Unnatural instructions: Tuning language models with (almost) no human labor.
\newblock In Anna Rogers, Jordan Boyd-Graber, and Naoaki Okazaki, editors, \emph{Proceedings of the 61st Annual Meeting of the Association for Computational Linguistics (Volume 1: Long Papers)}, pages 14409--14428, Toronto, Canada, July 2023. Association for Computational Linguistics.
\newblock \doi{10.18653/v1/2023.acl-long.806}.
\newblock URL \url{https://aclanthology.org/2023.acl-long.806}.

\bibitem[Kim et~al.(2023)Kim, Shin, Cho, Jang, Longpre, Lee, Yun, Shin, Kim, Thorne, et~al.]{kim2023prometheus}
Seungone Kim, Jamin Shin, Yejin Cho, Joel Jang, Shayne Longpre, Hwaran Lee, Sangdoo Yun, Seongjin Shin, Sungdong Kim, James Thorne, et~al.
\newblock Prometheus: Inducing fine-grained evaluation capability in language models.
\newblock \emph{arXiv preprint arXiv:2310.08491}, 2023.

\bibitem[K{\"o}pf et~al.(2023)K{\"o}pf, Kilcher, von R{\"u}tte, Anagnostidis, Tam, Stevens, Barhoum, Duc, Stanley, Nagyfi, et~al.]{kopf2023openassistant}
Andreas K{\"o}pf, Yannic Kilcher, Dimitri von R{\"u}tte, Sotiris Anagnostidis, Zhi-Rui Tam, Keith Stevens, Abdullah Barhoum, Nguyen~Minh Duc, Oliver Stanley, Rich{\'a}rd Nagyfi, et~al.
\newblock {OpenAssistant} conversations--democratizing large language model alignment.
\newblock \emph{arXiv preprint arXiv:2304.07327}, 2023.

\bibitem[Kwiatkowski et~al.(2019)Kwiatkowski, Palomaki, Redfield, Collins, Parikh, Alberti, Epstein, Polosukhin, Kelcey, Devlin, Lee, Toutanova, Jones, Chang, Dai, Uszkoreit, Le, and Petrov]{47761}
Tom Kwiatkowski, Jennimaria Palomaki, Olivia Redfield, Michael Collins, Ankur Parikh, Chris Alberti, Danielle Epstein, Illia Polosukhin, Matthew Kelcey, Jacob Devlin, Kenton Lee, Kristina~N. Toutanova, Llion Jones, Ming-Wei Chang, Andrew Dai, Jakob Uszkoreit, Quoc Le, and Slav Petrov.
\newblock Natural questions: a benchmark for question answering research.
\newblock \emph{Transactions of the Association of Computational Linguistics}, 2019.

\bibitem[Lee et~al.(2023)Lee, Phatale, Mansoor, Lu, Mesnard, Bishop, Carbune, and Rastogi]{lee2023rlaif}
Harrison Lee, Samrat Phatale, Hassan Mansoor, Kellie Lu, Thomas Mesnard, Colton Bishop, Victor Carbune, and Abhinav Rastogi.
\newblock {RLAIF}: Scaling reinforcement learning from human feedback with ai feedback.
\newblock \emph{arXiv preprint arXiv:2309.00267}, 2023.

\bibitem[Li et~al.(2024)Li, Yu, Zhou, Schick, Zettlemoyer, Levy, Weston, and Lewis]{li2023self}
Xian Li, Ping Yu, Chunting Zhou, Timo Schick, Luke Zettlemoyer, Omer Levy, Jason Weston, and Mike Lewis.
\newblock Self-alignment with instruction backtranslation.
\newblock In \emph{The Twelfth International Conference on Learning Representations}, 2024.
\newblock URL \url{https://openreview.net/forum?id=1oijHJBRsT}.

\bibitem[Li et~al.(2023)Li, Zhang, Dubois, Taori, Gulrajani, Guestrin, Liang, and Hashimoto]{alpaca_eval}
Xuechen Li, Tianyi Zhang, Yann Dubois, Rohan Taori, Ishaan Gulrajani, Carlos Guestrin, Percy Liang, and Tatsunori~B. Hashimoto.
\newblock Alpacaeval: An automatic evaluator of instruction-following models.
\newblock \url{https://github.com/tatsu-lab/alpaca_eval}, 2023.

\bibitem[Lin(2004)]{lin-2004-rouge}
Chin-Yew Lin.
\newblock {ROUGE}: A package for automatic evaluation of summaries.
\newblock In \emph{Text Summarization Branches Out}, pages 74--81, Barcelona, Spain, July 2004. Association for Computational Linguistics.
\newblock URL \url{https://aclanthology.org/W04-1013}.

\bibitem[Mihaylov et~al.(2018)Mihaylov, Clark, Khot, and Sabharwal]{OpenBookQA2018}
Todor Mihaylov, Peter Clark, Tushar Khot, and Ashish Sabharwal.
\newblock Can a suit of armor conduct electricity? a new dataset for open book question answering.
\newblock In \emph{EMNLP}, 2018.

\bibitem[Ouyang et~al.(2022)Ouyang, Wu, Jiang, Almeida, Wainwright, Mishkin, Zhang, Agarwal, Slama, Ray, et~al.]{ouyang2022training}
Long Ouyang, Jeffrey Wu, Xu~Jiang, Diogo Almeida, Carroll Wainwright, Pamela Mishkin, Chong Zhang, Sandhini Agarwal, Katarina Slama, Alex Ray, et~al.
\newblock Training language models to follow instructions with human feedback.
\newblock \emph{Advances in Neural Information Processing Systems}, 35:\penalty0 27730--27744, 2022.

\bibitem[Pan et~al.(2023)Pan, Saxon, Xu, Nathani, Wang, and Wang]{pan2023automatically}
Liangming Pan, Michael Saxon, Wenda Xu, Deepak Nathani, Xinyi Wang, and William~Yang Wang.
\newblock Automatically correcting large language models: Surveying the landscape of diverse self-correction strategies.
\newblock \emph{arXiv preprint arXiv:2308.03188}, 2023.

\bibitem[Radford et~al.(2019)Radford, Wu, Child, Luan, Amodei, Sutskever, et~al.]{radford2019language}
Alec Radford, Jeffrey Wu, Rewon Child, David Luan, Dario Amodei, Ilya Sutskever, et~al.
\newblock Language models are unsupervised multitask learners.
\newblock \emph{OpenAI blog}, 1\penalty0 (8):\penalty0 9, 2019.

\bibitem[Rafailov et~al.(2023)Rafailov, Sharma, Mitchell, Manning, Ermon, and Finn]{rafailov2023direct}
Rafael Rafailov, Archit Sharma, Eric Mitchell, Christopher~D Manning, Stefano Ermon, and Chelsea Finn.
\newblock Direct preference optimization: Your language model is secretly a reward model.
\newblock In \emph{Thirty-seventh Conference on Neural Information Processing Systems}, 2023.
\newblock URL \url{https://openreview.net/forum?id=HPuSIXJaa9}.

\bibitem[Saha et~al.(2023)Saha, Levy, Celikyilmaz, Bansal, Weston, and Li]{saha2023branch}
Swarnadeep Saha, Omer Levy, Asli Celikyilmaz, Mohit Bansal, Jason Weston, and Xian Li.
\newblock Branch-solve-merge improves large language model evaluation and generation.
\newblock \emph{arXiv preprint arXiv:2310.15123}, 2023.

\bibitem[Sap et~al.(2019)Sap, Rashkin, Chen, Bras, and Choi]{DBLP:journals/corr/abs-1904-09728}
Maarten Sap, Hannah Rashkin, Derek Chen, Ronan~Le Bras, and Yejin Choi.
\newblock Socialiqa: Commonsense reasoning about social interactions.
\newblock \emph{CoRR}, abs/1904.09728, 2019.
\newblock URL \url{http://arxiv.org/abs/1904.09728}.

\bibitem[Schulman et~al.(2017)Schulman, Wolski, Dhariwal, Radford, and Klimov]{schulman2017proximal}
John Schulman, Filip Wolski, Prafulla Dhariwal, Alec Radford, and Oleg Klimov.
\newblock Proximal policy optimization algorithms.
\newblock \emph{arXiv preprint arXiv:1707.06347}, 2017.

\bibitem[Stiennon et~al.(2020)Stiennon, Ouyang, Wu, Ziegler, Lowe, Voss, Radford, Amodei, and Christiano]{stiennon2020learning}
Nisan Stiennon, Long Ouyang, Jeffrey Wu, Daniel Ziegler, Ryan Lowe, Chelsea Voss, Alec Radford, Dario Amodei, and Paul~F Christiano.
\newblock Learning to summarize with human feedback.
\newblock \emph{Advances in Neural Information Processing Systems}, 33:\penalty0 3008--3021, 2020.

\bibitem[Taori et~al.(2023)Taori, Gulrajani, Zhang, Dubois, Li, Guestrin, Liang, and Hashimoto]{alpaca}
Rohan Taori, Ishaan Gulrajani, Tianyi Zhang, Yann Dubois, Xuechen Li, Carlos Guestrin, Percy Liang, and Tatsunori~B. Hashimoto.
\newblock Stanford alpaca: An instruction-following llama model.
\newblock \url{https://github.com/tatsu-lab/stanford_alpaca}, 2023.

\bibitem[Touvron et~al.(2023)Touvron, Martin, Stone, Albert, Almahairi, Babaei, Bashlykov, Batra, Bhargava, Bhosale, et~al.]{touvron2023llama2}
Hugo Touvron, Louis Martin, Kevin Stone, Peter Albert, Amjad Almahairi, Yasmine Babaei, Nikolay Bashlykov, Soumya Batra, Prajjwal Bhargava, Shruti Bhosale, et~al.
\newblock Llama 2: Open foundation and fine-tuned chat models.
\newblock \emph{arXiv preprint arXiv:2307.09288}, 2023.

\bibitem[Van~der Maaten and Hinton(2008)]{van2008visualizing}
Laurens Van~der Maaten and Geoffrey Hinton.
\newblock Visualizing data using {t-SNE}.
\newblock \emph{Journal of machine learning research}, 9\penalty0 (11), 2008.

\bibitem[Wang et~al.(2023)Wang, Kordi, Mishra, Liu, Smith, Khashabi, and Hajishirzi]{wang2022self}
Yizhong Wang, Yeganeh Kordi, Swaroop Mishra, Alisa Liu, Noah~A. Smith, Daniel Khashabi, and Hannaneh Hajishirzi.
\newblock Self-instruct: Aligning language models with self-generated instructions.
\newblock In Anna Rogers, Jordan Boyd-Graber, and Naoaki Okazaki, editors, \emph{Proceedings of the 61st Annual Meeting of the Association for Computational Linguistics (Volume 1: Long Papers)}, pages 13484--13508, Toronto, Canada, July 2023. Association for Computational Linguistics.
\newblock \doi{10.18653/v1/2023.acl-long.754}.
\newblock URL \url{https://aclanthology.org/2023.acl-long.754}.

\bibitem[Xiong et~al.(2023)Xiong, Dong, Ye, Zhong, Jiang, and Zhang]{xiong2023gibbs}
Wei Xiong, Hanze Dong, Chenlu Ye, Han Zhong, Nan Jiang, and Tong Zhang.
\newblock Gibbs sampling from human feedback: A provable kl-constrained framework for rlhf.
\newblock \emph{arXiv preprint arXiv:2312.11456}, 2023.

\bibitem[Xu et~al.(2023)Xu, Lee, Sukhbaatar, and Weston]{xu2023some}
Jing Xu, Andrew Lee, Sainbayar Sukhbaatar, and Jason Weston.
\newblock Some things are more cringe than others: Preference optimization with the pairwise cringe loss.
\newblock \emph{arXiv preprint arXiv:2312.16682}, 2023.

\bibitem[Yuan et~al.(2023)Yuan, Yuan, Tan, Wang, Huang, and Huang]{yuan2023rrhf}
Hongyi Yuan, Zheng Yuan, Chuanqi Tan, Wei Wang, Songfang Huang, and Fei Huang.
\newblock {RRHF}: Rank responses to align language models with human feedback.
\newblock In \emph{Thirty-seventh Conference on Neural Information Processing Systems}, 2023.
\newblock URL \url{https://openreview.net/forum?id=EdIGMCHk4l}.

\bibitem[Zellers et~al.(2019)Zellers, Holtzman, Bisk, Farhadi, and Choi]{DBLP:conf/acl/ZellersHBFC19}
Rowan Zellers, Ari Holtzman, Yonatan Bisk, Ali Farhadi, and Yejin Choi.
\newblock Hellaswag: Can a machine really finish your sentence?
\newblock In Anna Korhonen, David~R. Traum, and Llu{\'{\i}}s M{\`{a}}rquez, editors, \emph{Proceedings of the 57th Conference of the Association for Computational Linguistics, {ACL} 2019, Florence, Italy, July 28- August 2, 2019, Volume 1: Long Papers}, pages 4791--4800. Association for Computational Linguistics, 2019.
\newblock \doi{10.18653/V1/P19-1472}.
\newblock URL \url{https://doi.org/10.18653/v1/p19-1472}.

\bibitem[Zhao et~al.(2023)Zhao, Joshi, Liu, Khalman, Saleh, and Liu]{zhao2023slic}
Yao Zhao, Rishabh Joshi, Tianqi Liu, Misha Khalman, Mohammad Saleh, and Peter~J Liu.
\newblock {SLiC-HF}: Sequence likelihood calibration with human feedback.
\newblock \emph{arXiv preprint arXiv:2305.10425}, 2023.

\bibitem[Zheng et~al.(2023{\natexlab{a}})Zheng, Ke, Zhang, and Huang]{zheng2023click}
Chujie Zheng, Pei Ke, Zheng Zhang, and Minlie Huang.
\newblock Click: Controllable text generation with sequence likelihood contrastive learning.
\newblock In Anna Rogers, Jordan Boyd-Graber, and Naoaki Okazaki, editors, \emph{Findings of the Association for Computational Linguistics: ACL 2023}, pages 1022--1040, Toronto, Canada, July 2023{\natexlab{a}}. Association for Computational Linguistics.
\newblock \doi{10.18653/v1/2023.findings-acl.65}.
\newblock URL \url{https://aclanthology.org/2023.findings-acl.65}.

\bibitem[Zheng et~al.(2023{\natexlab{b}})Zheng, Chiang, Sheng, Zhuang, Wu, Zhuang, Lin, Li, Li, Xing, Zhang, Gonzalez, and Stoica]{zheng2023judging}
Lianmin Zheng, Wei-Lin Chiang, Ying Sheng, Siyuan Zhuang, Zhanghao Wu, Yonghao Zhuang, Zi~Lin, Zhuohan Li, Dacheng Li, Eric Xing, Hao Zhang, Joseph~E. Gonzalez, and Ion Stoica.
\newblock Judging {LLM}-as-a-judge with {MT}-bench and chatbot arena.
\newblock In \emph{Thirty-seventh Conference on Neural Information Processing Systems Datasets and Benchmarks Track}, 2023{\natexlab{b}}.
\newblock URL \url{https://openreview.net/forum?id=uccHPGDlao}.

\bibitem[Ziegler et~al.(2019)Ziegler, Stiennon, Wu, Brown, Radford, Amodei, Christiano, and Irving]{ziegler2019fine}
Daniel~M Ziegler, Nisan Stiennon, Jeffrey Wu, Tom~B Brown, Alec Radford, Dario Amodei, Paul Christiano, and Geoffrey Irving.
\newblock Fine-tuning language models from human preferences.
\newblock \emph{arXiv preprint arXiv:1909.08593}, 2019.

\end{thebibliography}
\bibliographystyle{plainnat}

\newpage
\appendix
\onecolumn

\section{Appendix}
\subsection{Distributions of IFT, EFT and AIFT data}
\label{app:ift_eft_distribution}

\begin{figure}[h]
    \centering
    \begin{subfigure}[b]{0.49\textwidth}
        \includegraphics[width=\textwidth]{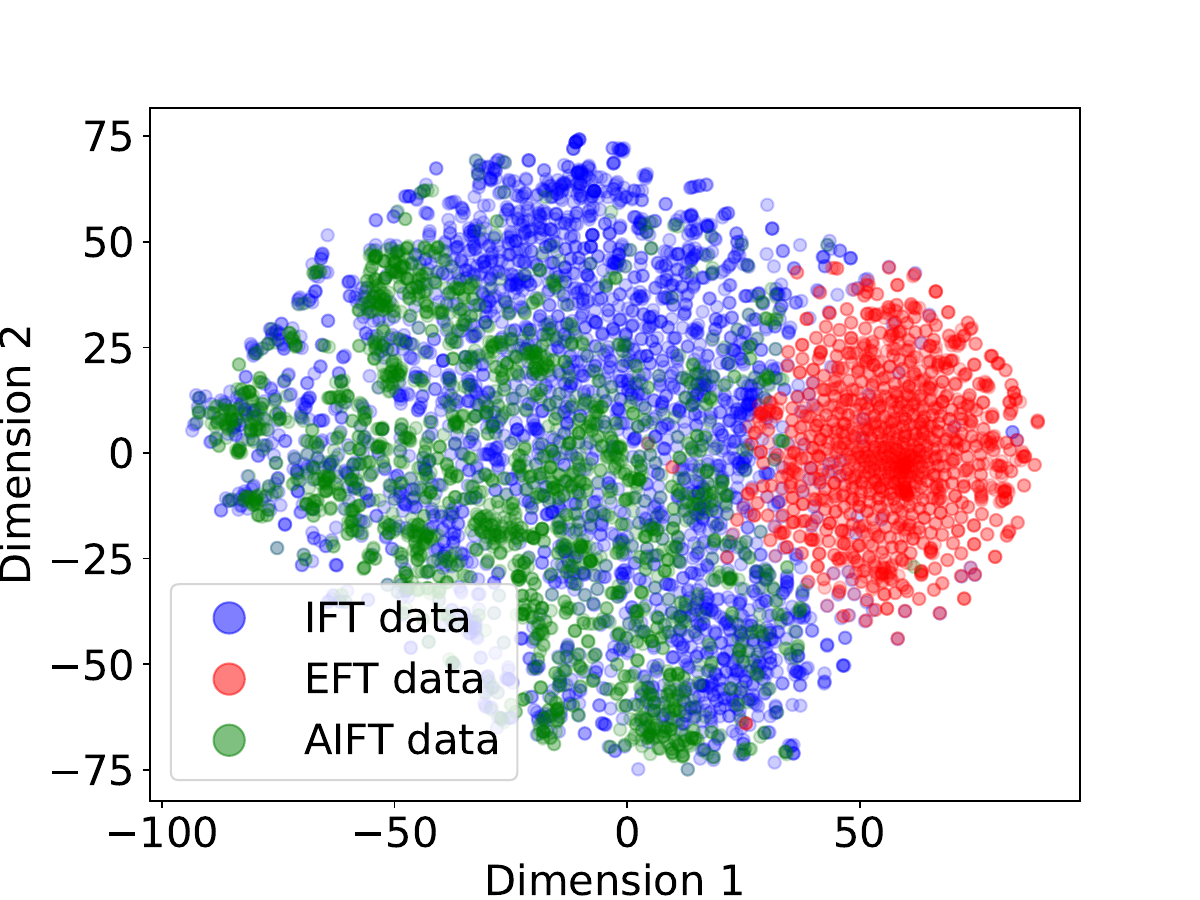}
        \caption{Instruction distribution of IFT, EFT and AIFT data.}
        \label{fig:instruction_dist}
    \end{subfigure}
    \begin{subfigure}[b]{0.49\textwidth}
        \includegraphics[width=\textwidth]{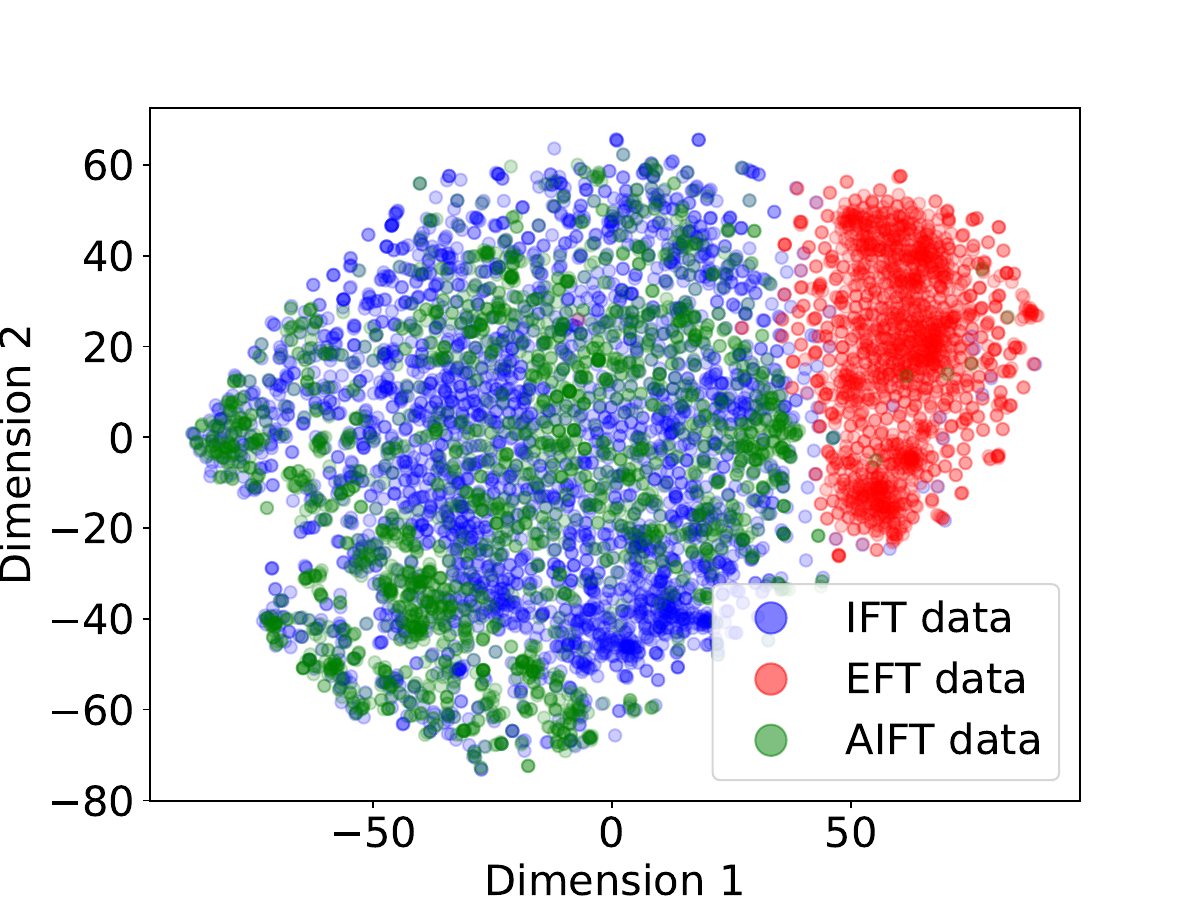}
        \caption{Response distribution of IFT, EFT, and AIFT data.}
        \label{fig:response_dict}
    \end{subfigure}
    \caption{Distributions of both instructions and responses for IFT, EFT  and AIFT data.}
    \label{fig:ift_eft_distribution}
\end{figure}

We have plotted the distribution of instructions for IFT, EFT and AIFT($M_1$) data, and the distribution of responses for IFT, EFT and AIFT($M_1$) data in \autoref{fig:ift_eft_distribution}. It is clear that the IFT data and EFT data come from very different distributions while the IFT and AIFT($M_1$) data come from similar distributions.

\subsection{EFT Prompts}
\label{app:xian_eft_prompt}

The EFT prompt which we use in our main experiments is shown in \autoref{tab:eval_prompt}. 

\paragraph{Other EFT prompts we have tried}

At first, we took the EFT prompt from \citet{li2023self} as shown in \autoref{tab:eval_prompt_xian}. However, we found that this prompt was not as effective as our additive score-counting prompt because the model needed to treat the task as a multiple-choice problem, and it was difficult for the model to break down this multiple-choice problem into sub-problems involving evaluating various aspects of the response. When using the model trained on 3,200 IFT data only, its performance on the EFT test set using our additive score-counting prompt and prompt from \citet{li2023self} is shown in \autoref{tab:compare_eft_prompt}.

\begin{figure}[t]
\centering
\begin{tcolorbox}[colback=green!10!white, 
                  colframe=green!30!white, 
                  width=0.99\textwidth, 
                  arc=4mm, 
                  auto outer arc,
                  ]
Below is a question from an user and a candidate response. Please grade the response on a 5-point scale using the following criteria:\\
\\
1: It means the answer is incomplete, vague, off-topic, controversial, or not exactly what the user asked for. For example, some content seems missing, numbered list does not start from the beginning, the opening sentence repeats user's question. Or the response is from another person’s perspective with their personal experience (e.g. taken from blog posts), or looks like an answer from a forum. Or it contains promotional text, navigation text, or other irrelevant information. \\
2: It means the answer addresses most of the asks from the user. It does not directly address the user's question. For example, it only provides a high-level methodology instead of the exact solution to user's question. \\
3: It means the answer is helpful but not written by an AI Assistant. It addresses all the basic asks from the user. It is complete and self contained with the drawback that the response is not written from an AI assistant's perspective, but from other people's perspective. The content looks like an excerpt from a blog post, web page, or web search results. For example, it contains personal experience or opinion, mentions comments section, or share on social media, etc.\\
4: It means the answer is written from an AI assistant's perspective with a clear focus of addressing the instruction. It provide a complete, clear, and comprehensive response to user’s question or instruction without missing or irrelevant information. It is well organized, self-contained, and written in a helpful tone. It has minor room for improvement, e.g. more concise and focused.\\
5: It means it is a perfect answer from an AI Assistant. It has a clear focus on being a helpful AI Assistant, where the response looks like intentionally written to address the user's question or instruction without any irrelevant sentences. The answer provides high quality content, demonstrating expert knowledge in the area, is very well written, logical, easy-to-follow, engaging and insightful.\\
\\
\\
User: \texttt{\color{red}<INSTRUCTION\_HERE>}\\
\\
\texttt{<}response\texttt{>}\texttt{\color{red}<RESPONSE\_HERE>}\texttt{<}/response\texttt{>}\\
\\
Please first briefly describe your reasoning (in less than 100 words), and then write ``Score: \texttt{<}rating\texttt{>}'' in the last line. Answer in the style of an AI Assistant, with knowledge from web search if needed. To derive the final score based on the criteria, let's think step-by-step.
\end{tcolorbox}
\caption{LLM-as-a-Judge prompt taken from \citet{li2023self}.}
\label{tab:eval_prompt_xian}
\end{figure}

\begin{table}[h]
\begin{center}
\begin{small}
\begin{sc}
  \setlength\tabcolsep{20pt}
\begin{tabular}{lcc}

\toprule

EFT Prompt             &   Multiple Choice prompt   &  Ours \\
\midrule
Pairwise accuracy $(\uparrow)$ & 26.6\% & 65.1\%\\
5-best \%    $(\uparrow)$ &  23.5\%  & 39.6\% \\
Exact Match \%  $(\uparrow)$ &  1.1\% & 10.1\% \\
Spearman corr. $(\uparrow)$ & -0.18 & 0.25 \\
Kendall $\tau$ corr.    $(\uparrow)$ & -0.16 & 0.23 \\
\bottomrule
\end{tabular}
\end{sc}
\end{small}
\end{center}
\caption{We tried various LLM-as-Judge prompts using the model trained with 3,200 IFT data only and found that our additive score-counting prompt worked best which demonstrates significant improvements in EFT performance comparing to the prompt used by \citet{li2023self}. 
\label{tab:compare_eft_prompt}
}
\end{table}
\subsection{Self-rewarding Models Using IFT Data Only}
\label{app:self_reward_ift_only}
To demonstrate the importance of the EFT data, we also trained a series of models starting with the model trained only on the IFT data. The following is the model sequence.

\begin{itemize}
\item[$M_0$]: Base pretrained LLM with no fine-tuning.
\item[$M_1^\prime$]: Initialized with $M_0$, then fine-tuned on the IFT seed data only using SFT.
\item[$M_2^\prime$]: Initialized with $M_1^\prime$, then trained with AIFT($M_1^\prime$) data using DPO.
\item[$M_3^\prime$]: Initialized with $M_2^\prime$, then trained with AIFT($M_2^\prime$) data using DPO.
\end{itemize}

\begin{figure*}[t]
    \centering
     \includegraphics[width=0.7\textwidth,trim={0 -1cm 0  0}]{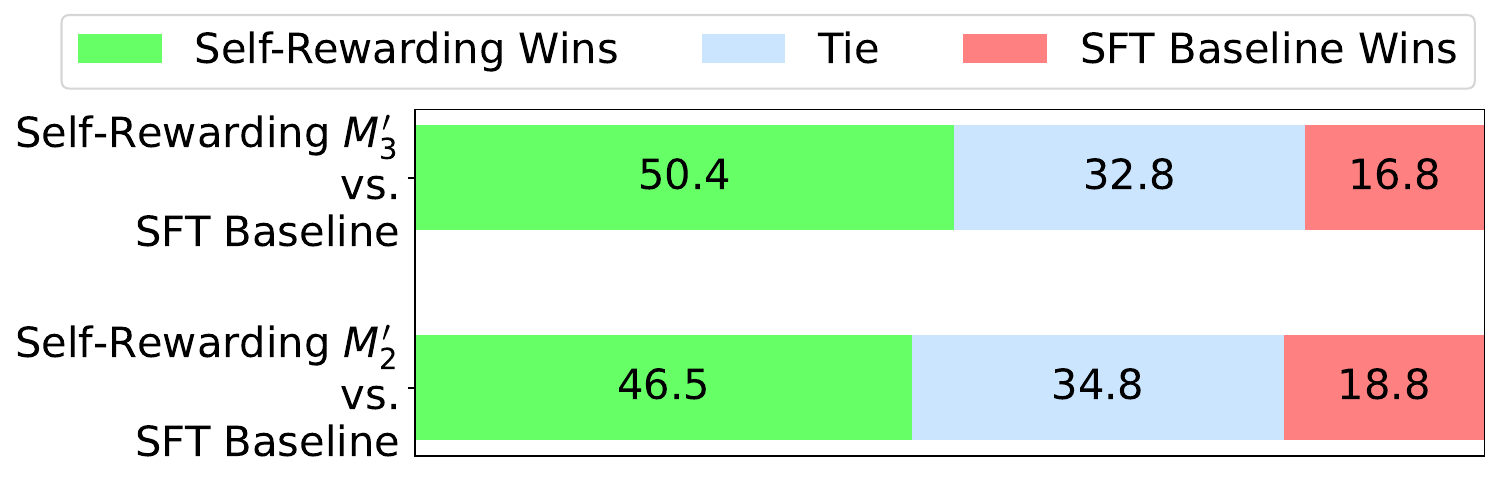}
        \includegraphics[width=0.7\columnwidth]{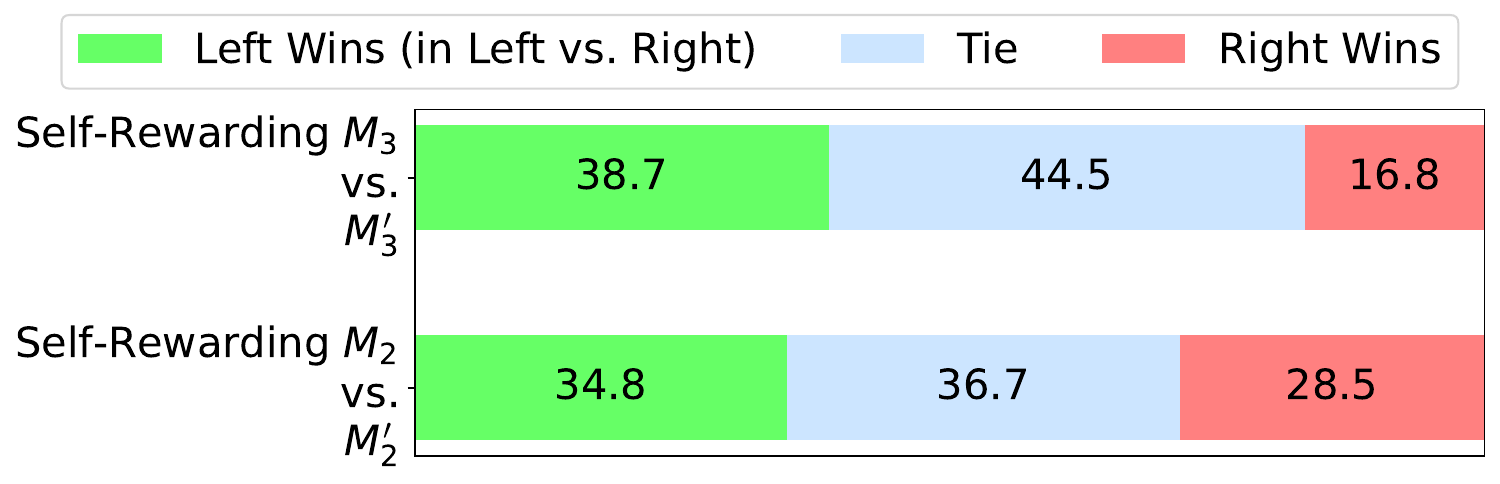}
    \caption{\textbf{EFT data helps the self-rewarding loop:} We evaluated the series of models trained using self-reward loops starting from the model trained using only IFT data. We performed head-to-head win rates comparisons on the IFT test set. While $M_2^\prime$ can improve over the SFT baseline and $M_3^\prime$ can improve even more over the SFT baseline, they lag far behind the corresponding models ($M_2$, $M_3$) that started from a base model trained using both IFT and EFT data, see \autoref{fig:h2h}.}
    \label{fig:self_reward_from_sft_baseline}
\end{figure*}

Since we did not use EFT data to train the series of models, they were not always able to score the responses according to the format and even when they did, the scores given typically converged to 4. Therefore, even when starting from the same number of generated new prompts, we could only collect a very small number of valid training samples for DPO.
In total, we collected 541 pairs to form the AIFT($M_1^\prime$) dataset used to train $M_2^\prime$ via DPO, and 429 pairs to form AIFT($M_2^\prime$) used to train $M_3^\prime$. The win rates are shown in \autoref{fig:self_reward_from_sft_baseline}. From the figure we can conclude that EFT data helps to get better performance in the same number of iterations and the gap in performance between the model trained with EFT data and the model trained without EFT data widens in the later iterations.

\begin{figure}[t]
\centering
\small
\begin{tcolorbox}[colback=green!10!white, 
                  colframe=green!30!white, 
                  width=0.99\textwidth, 
                  arc=4mm, 
                  auto outer arc,
                  ]
\texttt{\color{red}<LIST ALL ALPACAEVAL INSTRUCTIONS>}\\                  
Given the above list of possible instructions, define  a maximum of 20 categories that would cover the types of intructions, for example recipes, reasoning tasks, general knowledge etc. Try to cover as many of the instructions as possible with the maximum 20 categories, while keeping the categories high-level, simple and easy to understand.
\end{tcolorbox}
\caption{Prompt used to obtain instruction categories on the AlpacaEval test set.}
\label{tab:instruction_categories}
\end{figure}

\begin{figure}[th!]
\centering
\small
\begin{tcolorbox}[colback=green!10!white, 
                  colframe=green!30!white, 
                  width=0.99\textwidth, 
                  arc=4mm, 
                  auto outer arc,
                  ]
Instruction: \texttt{\color{red}<INSTRUCTION>}

Given the above, categorize it into one of the following 20 categories:\\
\texttt{\color{red}<LIST ALL CATEGORIES>}\\
\\
Secondly, score the instruction in terms of complexity: how complex you think it is to answer from 1-10 (where 10 is a complex question whereby first reasoning or breaking down the question into multiple subquestions for example might help improve the answer).\\
\\
Thirdly, indicate how long you think the response to the instruction should be, either (a) 1 sentence, (b) 1-3 sentences, (c) 1 paragraph, (d) 2 paragraphs, or (e) 3 or more paragraphs.\\
\\
\\
Provide your final response in the following format:\\
Category: \texttt{<}one of the 20 categories\texttt{>}\\
Complexity: \texttt{<}score out of 10\texttt{>}\\
Length: \texttt{<}length category\texttt{>}. Do not provide the actual response. 
\end{tcolorbox}
\caption{Prompt for categorizing instructions based on their topics, complexities and expected response lengths.}
\label{tab:instruction_category_complexity_expected_length_prompt}
\end{figure}

\begin{table}[t]
\centering
   \caption{
   \label{tab:instruction_category_breakdown} Breakdown of AlpacaEval test set instructions by instruction category.
   }
\begin{tabular}{lrr}
\toprule
\textbf{Category}     & \textbf{Number} & \textbf{Percentage}  \\
\midrule
Science / Technology / Engineering                  & 134    & 16.65\%    \\
Professional / Business / Marketing                 & 77     & 9.57\%      \\
Social Interaction / Relationships / Human Behavior & 68     & 8.45\%      \\
Miscellaneous / Other                               & 61     & 7.58\%      \\
Mathematics / Logical Reasoning                     & 52     & 6.46\%      \\
Cooking / Recipes                                   & 48     & 5.96\%      \\
Software Development / Coding / Algorithms          & 44     & 5.47\%      \\
Travel / Geography / Exploration                    & 41     & 5.09\%      \\
Literature / Writing / Communication                & 39     & 4.84\%      \\
History / Social Studies                            & 38     & 4.72\%      \\
Entertainment / Media Analysis                      & 34     & 4.22\%      \\
Language Learning / Linguistics                     & 32     & 3.98\%      \\
Music / Audio / Arts                                & 30     & 3.73\%      \\
DIY Projects / Hobbies                              & 24     & 2.98\%      \\
Technology / Gadgets / Consumer Products            & 20     & 2.48\%      \\
Gaming / Game Development                           & 18     & 2.24\%      \\
Exercise / Health / Wellness                        & 16     & 1.99\%      \\
Philosophy / Ethics / Ideology                      & 15     & 1.86\%      \\
Sports / Athletics / Physical Activity              & 12     & 1.49\%      \\
Strategy / Problem-Solving / Critical Thinking      & 2      & 0.24\%      \\
\bottomrule
\end{tabular}
\end{table}

\begin{table}[t]
\centering
   \caption{
   \label{tab:instruction_complexity_breakdown} Breakdown of AlpacaEval test set instructions by instruction complexity. The instructions increase in complexity from 1 to 9, where 10 is a complex question that requires first reasoning or breaking the problem into sub-problems before it can be solved.
   }
\begin{tabular}{lrr}
\toprule
\textbf{Complexity}     & \textbf{Number} & \textbf{Percentage}  \\
\midrule
3 &  238 & 29.57\% \\
2 & 206 & 25.59\% \\
4 & 122 & 15.16\% \\
6 & 79 & 9.81\% \\
5 & 68 & 8.45\% \\
7 & 41 & 5.09\% \\
1 & 34 & 4.22\% \\
8 & 14 & 1.74\% \\
9 & 3 & 0.37\% \\
\bottomrule
\end{tabular}
\end{table}

\begin{table}[t]
\centering
   \caption{
\label{tab:instruction_expected_response_length_breakdown} Breakdown of AlpacaEval test set instructions by expected response length.
   }
\begin{tabular}{lrr}
\toprule
\textbf{Expected Length}     & \textbf{Number} & \textbf{Percentage}  \\
\midrule
1-3 sentences & 361 & 44.84\% \\
1 paragraph & 269 & 33.42\% \\
1 sentence & 143 & 17.76\% \\
2 paragraphs & 31 & 3.85\% \\
3 or more paragraphs & 1 & 0.13\% \\
\bottomrule
\end{tabular}
\end{table}

\begin{figure}[h!]
    \centering
    \hfill
    \begin{subfigure}[b]{0.47\linewidth}
        \includegraphics[width=\columnwidth]{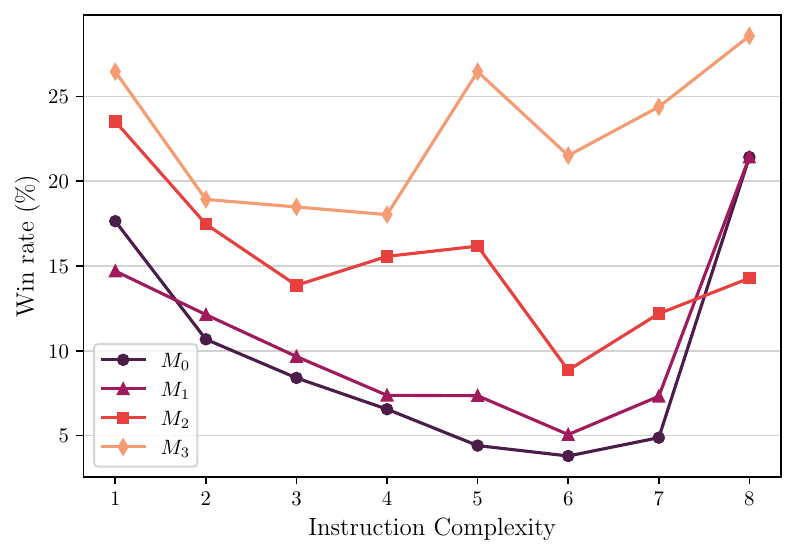}
        \label{fig:complexity_winrate}
    \end{subfigure}
    \hfill
     \begin{subfigure}[b]{0.485\linewidth}
        \includegraphics[width=\columnwidth]{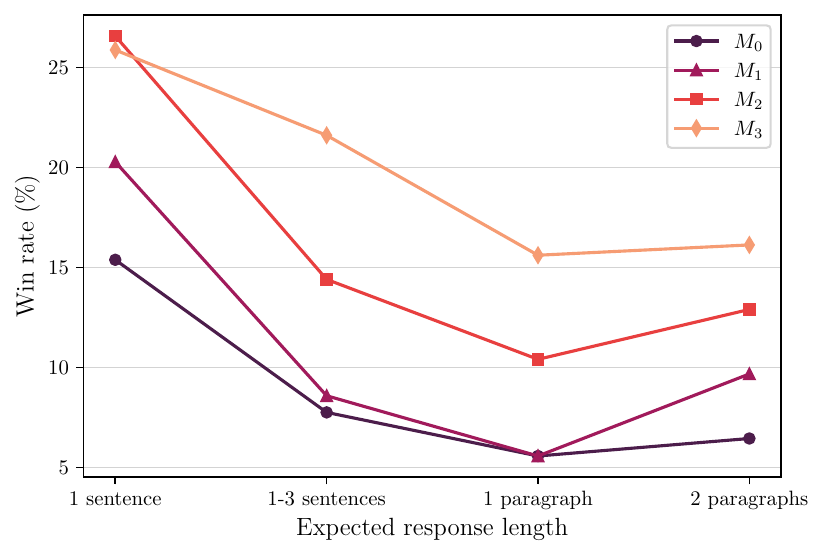}
        \label{fig:expected_response_len_winrate}
    \end{subfigure}
    \hfill
    \captionsetup{skip=0pt}  
    \caption{AlpacaEval win rate breakdown for instruction complexities (left) and expected response lengths (right).
    Self-Rewarding models give gains across most complexities and all response length ranges.
    }
    \label{fig:fine_grained_winrate_detailed_other_two}
\end{figure}

\subsection{Preference optimization outperforms augmenting with positive examples only}
\label{app:positive_examples}
We also tried an alternative self-training procedure
of adding high-quality self-instruction creation examples to supervised
fine-tuning (without preference optimization), rather than DPO. 
 In this variant, we add additional examples of (instruction prompt, response) curated by the model to the seed set for supervised fine-tuning, following other approaches \citep{li2023self,adolphs2022cringe,gulcehre2023reinforced}, rather than constructing preference data. In this setup we only add examples where the candidate response was evaluated to give a perfect score of $r_i^n=5$.  Unfortunately we could not find a configuration where this approach helped. For example, adding 11,254 such examples that scored 5 out of 5, and optimizing the mixing weight in training, still yielded a head to head with the SFT Baseline of 29\% wins vs 30\% wins, i.e., no improvement.

\subsection{Augmented Prompt Generation Using Newly Trained Models}
\label{app:augmented_prompts_generation_using_newly_trained_models}

In our experiments, for time efficiency, we have created a fixed pool of augmented prompts in advance using ChatLlama 70B. In a real interactive system, ideally, those prompts could come from real users so that we can ensure the models are trained to align with real user requirements. Here, we also examine whether  our newly trained Self-Rewarding models in each iteration can generate new prompts through in-context learning, instead of using ChatLlama 70B.
To check this, we constructed 30 prompts with in-context examples using the original seed IFT data as described in \autoref{sec:creation} and tested whether $M_1$, $M_2$ and $M_3$ still possess in-context learning ability and can generate high quality instructions. According to manual inspection, all models can generate novel instructions given in-context examples in all 30 cases. However, for $M2$ and $M3$, the model is likely to first generate a few instructions, then generate a separator, and then start responding to the instructions, so some postprocessing might be necessary.

\subsection{AlpacaEval Test Sample Clustering}
\label{app:alpacaeval_test_sample_clustering}
We used the GPT-4 (\texttt{gpt-4-1106-preview}) model to categorize the instructions in the AlpacaEval test set into clusters from three perspectives: (1) instruction category, (2) instruction complexity, and (3) expected response length. To obtain instruction categories for the AlpaceEval test set, we used the prompt in \autoref{tab:instruction_categories} and obtained 20 categories in total. Then, to cluster the instructions into different groups, we use the prompt in \autoref{tab:instruction_category_complexity_expected_length_prompt} for each test example. The corresponding statistics are given in \autoref{tab:instruction_category_breakdown}, \autoref{tab:instruction_complexity_breakdown}, \autoref{tab:instruction_expected_response_length_breakdown}. The fine-grained results on instruction complexity and expected response length are given in \autoref{fig:fine_grained_winrate_detailed_other_two}.

\newpage

\begin{table}[h]
\small
\setlength{\tabcolsep}{3.9pt}
    \caption{
    {{\bf NLP Benchmarks}. Self-Rewarding models mostly tend to maintain performance compared to the Llama 2 base model and the SFT Baseline, despite being fine-tuned on very different instruction-following prompts.}}
    \vspace{2mm}
  \label{tab:core_knowledge_eval_results_detailed}
  \centering
  \scalebox{0.86}{
\begin{tabular}{lccccccccc}
\toprule
                & \multicolumn{5}{c}{{\textbf{Commonsense Reasoning}}}                                                                                                                                                    & \multicolumn{1}{c}{{\textbf{Math}}}                                       & \multicolumn{3}{c}{{\textbf{World Knowledge}}}                                                                                                                                                                                                                            \\
                     \cmidrule(lr){2-6}   \cmidrule(lr){7-7} \cmidrule(lr){8-10} 
                & \multicolumn{1}{c}{{\begin{tabular}[c]{@{}c@{}}ARC\\easy\end{tabular}}} 
                & \multicolumn{1}{c}{{\begin{tabular}[c]{@{}c@{}}ARC\\challenge\end{tabular}}}
                & \multicolumn{1}{c}{{HellaSwag}} & \multicolumn{1}{c}{{SIQA}} & \multicolumn{1}{c}{{PIQA}} & \multicolumn{1}{c}{{\begin{tabular}[c]{@{}c@{}}GSM8K\\ (em)\end{tabular}}} & \multicolumn{1}{c}{{\begin{tabular}[c]{@{}c@{}}MMLU\\ (macro\_avg/acc)\end{tabular}}} & \multicolumn{1}{c}{{\begin{tabular}[c]{@{}c@{}}OBQA\\ (acc\_comp)\end{tabular}}} & \multicolumn{1}{c}{{\begin{tabular}[c]{@{}c@{}}NQ\\ (em)\end{tabular}}} \\
                \midrule
{Llama 2} & 80.20                                   & 57.40                                        & 85.30                                   & 50.70                              & 82.80                              & 56.80                                                                              & 68.90                                                                                         & 60.20                                                                                    & 25.30                                                                           \\
{SFT Baseline}     & 76.49                                  & 55.97                                       & 85.17                                  & 51.48                             & 82.59                             & 50.72                                                                             & 69.76                                                                                  & 57.80                                                                                    & 34.35                                                                          \\
{$M_1$}     & 78.14                                  & 57.51                                       & 84.99                                  & 53.02                             & 82.92                             & 60.27                                                                             & 69.34                                                                                 & 57.60                                                                                    & 35.48                                                                          \\
{$M_2$}     & 74.84                                  & 54.51                                       & 84.27                                  & 51.23                             & 81.94                             & 59.29                                                                             & 69.31                                                                                  & 57.60                                                                                    & 33.07                                                                          \\
{$M_3$}     & 72.35                                  & 53.13                                       & 83.29                                  & 49.28                             & 80.79                             & 57.70                                                                              & 69.37                                                                                  & 58.40                                                                                    & 31.86  \\
\bottomrule
\end{tabular}
}
\end{table}
\begin{table}[h]
\footnotesize
\setlength{\tabcolsep}{4pt}
    \caption{
    {\textbf{MT-Bench Fine-grained Results}. We list our models' performance on each problem category. Self-reward is especially effective in improving the model's ability in writing, role-playing, extraction, and STEM tasks.}}
    \vspace{2mm}
  \label{tab:mt_bench_detailed}
  \centering
   \scalebox{0.88}{
\begin{tabular}{lccccccccc}
\toprule
             & \textbf{Writing} & \textbf{Roleplay} & \textbf{Reasoning} & \textbf{Math} & \textbf{Coding} & \textbf{Extraction}        & \textbf{STEM}  & \textbf{Humanities} & \textbf{Overall}          \\
\midrule
SFT  & 8.83   & 8.15     & 5.30       & 3.00  & 3.50    & 6.90 & 9.18 & 9.95       & 6.85 \\
M1           & 9.10     & 7.65     & 4.35      & 3.05 & 4.10    & 7.20               & 8.93 & 9.85       & 6.78         \\
M2           & 9.10     & 8.00      & 4.60       & 3.30  & 4.25   & 7.65              & 9.40   & 9.80        & 7.01           \\
M3           & 9.58   & 8.73    & 4.80       & 3.50  & 4.20    & 7.80               & 9.45  & 9.95       & 7.25           \\

\bottomrule
\end{tabular}
}
\end{table}

\subsection{NLP Benchmark Results and MT-Bench Results}
We provide the detailed model performance on a number of NLP benchmarks in \autoref{tab:core_knowledge_eval_results_detailed} and on MT-Bench in \autoref{tab:mt_bench_detailed}. In particular, some NLP benchmarks including ARC-Challenge, HellaSwag, SIQA, PIQA, and OBQA are all text completion tasks. In these tasks, given the multiple choice options, we choose the option corresponding to the highest log probability scored by the models as the final answer. As such, the objective of these particular tasks is quite different from what our algorithm tries to optimize, so the results on these tasks may not reflect the true capability of our models.

\end{document}